\documentclass[10pt,letterpaper]{IEEEtran}
\usepackage{graphicx,multicol,url,multirow,amssymb,color,booktabs}

\newcommand{\eqref}[1]{eq. (\ref{#1})}

\newcommand{\ignore}[1]{}

\makeatletter
\newcommand\footnoteref[1]{\protected@xdef\@thefnmark{\ref{#1}}\@footnotemark}
\makeatother

\newcommand\blfootnote[1]{%
  \begingroup
  \renewcommand\thefootnote{}\footnote{#1}%
  \addtocounter{footnote}{-1}%
  \endgroup
}

\begin{document}

\title{Video Processing from Electro-optical Sensors for Object Detection and Tracking in Maritime Environment: A Survey}

\author{Dilip K. Prasad$^{1,*}$, Deepu Rajan$^2$, Lily Rachmawati$^3$, Eshan Rajabally$^4$, and Chai Quek$^2$}
\maketitle
\blfootnote{\IEEEcompsocitemizethanks{\IEEEcompsocthanksitem{$^1$Rolls-Royce@ NTU Corporate Lab, Singapore $^*$dilipprasad@gmail.com, } \IEEEcompsocthanksitem{$^2$School of Computer Science and Engineering, Nanyang Technological University, Singapore}\IEEEcompsocthanksitem{$^3$Rolls-Royce Pvt. Ltd., Singapore}\IEEEcompsocthanksitem{$^4$Rolls Royce plc, Derby, United Kingdom}}}



\begin{abstract}
We present a survey on maritime object detection and tracking approaches, which are essential for the development of a navigational system for autonomous ships. The electro-optical (EO) sensor considered here is a video camera that operates in the visible or the infrared spectra, which conventionally complement radar and sonar and have demonstrated effectiveness for situational awareness at sea has demonstrated its effectiveness over the last few years. This paper provides a comprehensive overview of various approaches of video processing for object detection and tracking in the maritime environment. We follow an approach-based taxonomy wherein the advantages and limitations of each approach are compared.

The object detection system consists of the following modules: horizon detection, static background subtraction and foreground segmentation. Each of these has been studied extensively in maritime situations and has been shown to be challenging due to the presence of background motion especially due to waves and wakes. The main processes involved in object tracking include video frame registration, dynamic background subtraction, and the object tracking algorithm itself. The challenges for robust tracking arise due to camera motion, dynamic background and low contrast of tracked object, possibly due to environmental degradation. The survey also discusses multisensor approaches and commercial maritime systems that use EO sensors. The survey also highlights methods from computer vision research which hold promise to perform well in maritime EO data processing. Performance of several maritime and computer vision techniques is evaluated on newly proposed Singapore Marine Dataset.
\end{abstract}

\vspace{-2mm}
\section{Introduction}\label{sec:intro}

Maritime surveillance is a critical part of law enforcement and environment protection for littoral nations. However, with the growth of commercial ocean liners and other seafaring vessels such as cruise ships, technologies that have been traditionally deployed for military purposes, e.g. radars and sonars, are found to be of immense utility in providing support for navigation as well. The International Regulations for Preventing Collisions at Sea 1972 (COLREGs) requires all ships to be equipped with radars for proper lookout to provide early warning of potential collision. 
%
However, radar measurements are sensitive to the meteorological condition and the shape, size, and material of the targets. Thus, radar data has to be supplemented by other situational awareness sensors for better collision avoidance and navigation. 

Situational awareness at sea would undergo a paradigm shift with future development of the autonomous ship equipped with numerous sensors to support advanced decision and remote operation \cite{porathe2014situation}. Autonomy in ship navigation would lead to reduction in crew numbers as a result of re-skilling and relocation of crew to the shore, potentially resulting in less vigilant look-out. 
It is imperative that ranging devices are augmented with other sensors so that fail-safe decisions can be rapidly taken with high level of confidence.

\begin{table*}
  \centering
  \caption{Comparison of sensors used in maritime scenario for situation awareness.}\label{tab:sensors}
  \vspace{-2mm}
  \begin{tabular}{|l|p{1cm}|p{6.2cm}|p{7.4cm}|}
    \hline
    Sensor & Distance & \multicolumn{1}{|c|}{Advantages/Characteristics} & \multicolumn{1}{|c|}{Disadvantages} \\
    \hline
    \hline
    {} & & $\odot$ Long range sensing ability & $\otimes$ Needs separate systems for small range detections \\
    {Sonar} & $\sim$ 1 km & $\odot$ Underwater detection & $\otimes$ Performs poorly for objects with small acoustic  \\
    {\cite{elfes1987sonar}$-$\nocite{hansen2013synthetic,horne2000acoustic}\cite{hayes2009synthetic}} & to few  & $\odot$ Detects objects with large acoustic signatures  & $\quad$ signatures (ex. growler, small boats, and debris)\\
     & 100 km & $\quad$  (ex. whales and icebergs) & $\otimes$ Requires specialized user training\\
    \hline
    {} & & $\odot$ Long range sensing ability & $\otimes$ Suffers from minimum range  \\
    {Radar} & $\sim$ 1 km & $\odot$ Detects objects with high radar cross-sections & $\otimes$ Cannot penetrate water\\
    {\cite{ward1990maritimeradar}$-$\nocite{watts1990maritime,vicen2011ship,pasquariello1998automatic}\cite{ponsford2001integrated}} & to few & $\quad$ (mostly metallic) & $\otimes$ Cannot detect big objects with small radar cross-section \cite{szpak2011maritime}\\
     & 100 kms & {$\odot$ Large on-board power supply requirement} & $\otimes$ Requires specialized user training\\
    \hline
    {} & & $\odot$ Processes color information & $\otimes$ Sensitive to illumination and weather changes \\
    {} & $\sim$ m to & $\odot$ High resolution, advanced optics available& $\otimes$ Not suitable for night vision\\
    Visible range & $\sim$ km & $\odot$ Adaptive to new technology & $\otimes$ Computation intensive \\
    electro-optical& & $\odot$ Uses image processing/computer vision algorithms & $\otimes$ Low range sensing due to atmospheric attenuation\\
    {{\cite{szpak2011maritime}$-$\nocite{bloisi2009argos,bloisi2014background,bloisi2011automatic,fefilatyev2006horizon,fefilatyev2010tracking,van2009polynomial,hu2011robust,mittal2004motion,Ren2012,socek2005hybrid,strickland1997wavelet,sumimoto1994machine,wang2014aquatic,wang2015aquatic,wei2009automated,Zhou2014,gershikov2013horizon,bhanu1990model}} \cite{van2008discriminating}} & {} & $\odot$ Naturally intuitive, no need of user training & $\otimes$ Difficult to detect far objects and predict their size and distance \\
    {} & {} & {} & $\otimes$ Difficult to model water dynamics, wakes, and foam\\
    \hline
    {} & & $\odot$ Longer range than visible range EO  & $\otimes$ Significantly poorer optics available\\
    {} & $\sim$ m to & $\odot$ Allows night vision & $\otimes$ Saturated images in day time\\
    {Infrared range} & $\sim$ km & $\odot$ Water appears less dynamic& $\otimes$ Sensitive to illumination and weather changes\\
    {electro-optical} & & $\odot$ Intuitive, no need of user training& $\otimes$ Computation intensive\\
    {\cite{gershikov2013horizon}$-$\nocite{bhanu1990model,van2008discriminating,bouma2008automatic,van2000detection,van2014ship,van2014recognition,Chen2014,schwering2007eo,smith1999identification,tang2013research,tu2014infrared,Wang2006SVMWavelets,Wang2011FuzzyClutter}\cite{withagen1999automatic}} & {} & $\odot$ Adaptive to new technology & $\otimes$ Difficult to detect far objects and predict their size and distance   \\
    {} & {} & $\odot$ Uses image processing/computer vision algorithms & {$\otimes$ Horizon not-well defined in IR images} \\
    \hline
    \end{tabular}
  \vspace{-4mm}
\end{table*}

Electro-optical (EO) sensors are primed to complement ranging devices. In this paper, EO sensors imply video cameras operating in the visible and infrared portions of the electromagnetic spectrum. Some works \cite{bloisi2009argos,bloisi2015bookchapter} even recommend them as a replacement to ranging devices in special circumstances such as populated urban maritime scenario. EO sensors are of interest for two major reasons. Firstly, the image streams generated by them are directly interpretable and intuitive for human operators, alleviating the need for specialized training. Secondly, the image streams from them are amenable to image processing and computer vision such that advanced intelligence can be generated computationally without significant human intervention. Visible range EO sensors benefit from the availability of color data and high quality optics. On the other hand, infrared EO sensors benefit from night time visibility and suppression of highly dynamic regions in the video. This helps in the development of robust video processing algorithms \cite{robert2007multispectral}.

However, there are some disadvantages associated with EO data \cite{prasad2017challenges}. Although the atmospheric propagation characteristics for long wave infrared spectrum are superior to other visible and infrared frequencies \cite{chen1975attenuation}, in general, the atmospheric propagation losses restrict the range of the EO sensors to only a few kilometers. Further, EO data processing for automatic intelligence generation is quite challenging for maritime environment. Some of the challenges are:
\begin{itemize}
  \item the difficulty in modeling the dynamics of water (including waves, wakes and foams) for background subtraction and detection of foreground objects,
  \item variations in object appearances due to distance and angle of viewing, and 
  \item changes in illumination and weather conditions, such as due to clouds, sunshine, rain, glint, etc.
\end{itemize}

\begin{figure}
  \centering
  \includegraphics[width=\linewidth]{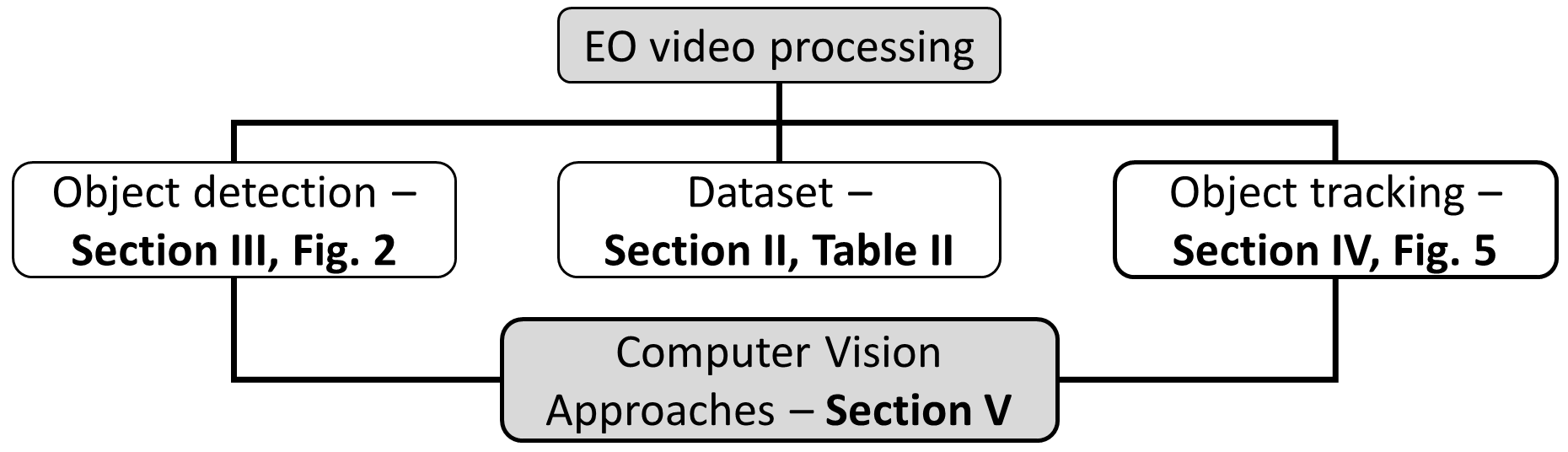}\vspace{-2mm}
  \caption{\vspace{-2mm}Organization of the survey.}\label{fig:overview}
  \vspace{-2mm}
\end{figure}

This paper presents a taxonomic survey of the approaches for processing EO data acquired from maritime environment. The organization of this survey is given in Fig. \ref{fig:overview}. Table \ref{tab:sensors} outlines the advantages and disadvantages of sonar, radar, and EO sensors. The survey focuses on maritime object detection and tracking using EO data to fulfil the navigational needs of an autonomous ship. The EO data is assumed to be available in the form of a video, either in the visible spectrum or in the infrared range. We exclude special cameras such as for monocular or stereovision from this survey. Survey on monocular and stereovision can be found in \cite{park2015passive,wang2013stereovision}. Further, we exclude device-level signal processing and high-level intelligence generation (such as vehicle behavior \cite{sivaraman2013looking}). We discuss post processing of the tracking data, maritime multi-sensor approaches, and commercial maritime systems that use EO sensors in Appendix.

\vspace{-2mm}
\section{Maritime dataset for comparative evaluation}\label{sec:dataset}

Works in maritime image processing typically use military owned or proprietary datasets which are not made available for research purposes. The authors are aware of only one dataset MarDCT \footnote{\url{http://www.dis.uniroma1.it/~labrococo/MAR/}} that is available online for academic and research purposes. Although this dataset does have images and videos acquired from both visible range and infrared range sensors, they are either in urban navigation scenario atypical of the usual maritime scenario or consider very simple scenarios with only one or two maritime vessels close to horizon.
There is a pressing need for a benchmark dataset of maritime videos so that quantitative comparison of various algorithms can be performed. To this end, we have created Singapore Marine Dataset, using Canon 70D cameras around Singapore waters. All the videos are acquired in high definition (1080 $\times$ 1920 pixels). We divide the dataset into parts, 32 on-shore videos and 4 on-board videos, which are acquired by camera placed on-shore on fixed platform and camera placed on-board a moving vessel, respectively.
Annotation tools developed in Matlab were used by volunteers not related to the project for annotation of ground truths (GTs) of horizon and objects in each frame. The dataset and annotation files of the GTs for horizon, objects, and tracks are available at the project webpage\footnote{\url{https://sites.google.com/site/dilipprasad/home/singapore-maritime-dataset}}. Details of the dataset are given in Table \ref{tab:dataset}.

\begin{table}
  \centering
  \caption{Details of the Singapore Marine Dataset.}\label{tab:dataset} \vspace{-2mm}
  \begin{tabular}{|p{3.5cm}||c|c|}
    \hline
    {} & On-board videos & On-shore videos \\
    \hline
    \hline
    Number of videos & 4 & 32 \\
    Total number of frames & 1196 & 16254 \\
    Number of frames in a video & $299$ & $[206,995]$\\
    Size of frames (pixels) & 1920 $\times$ 1080  & 1920 $\times$ 1080\\
    \hline
    \hline
    \multicolumn{3}{|c|}{Horizon and registration related}\\
    \hline
    $Y$ (pixels) & $[190.6,1077.1]$ &
    $[283.2, 925.6]$ \\
    Mean($Y$) $\pm$ standard & $552.5 \pm 183.9$ & $530.8 \pm 107.1$ \\
    deviation($Y$) (pixels) & & \\
    \hline
    $\alpha$ ($^\circ$) & $[-27.13, 0.40]$ & $[3.36,8.43]$ \\
    Mean($\alpha$) $\pm$ standard & $-7.18 \pm 5.80$ & $6.34 \pm 1.00$ \\
     deviation($\alpha$) ($^\circ$) & & \\
    \hline
    \hline
    \multicolumn{3}{|c|}{Object detection and tracking related}\\
    \hline
    \multicolumn{2}{|p{6cm}||}{Number of objects per frame, range} & $[2,20]$\\
    \multicolumn{2}{|p{6cm}||}{Number of stationary objects per frame, range} & $[0,14]$ \\
    \multicolumn{2}{|p{6cm}||}{Number of moving objects per frame} & $[0,10]$ \\
    \multicolumn{2}{|p{6cm}||}{Total number of object annotations in a video} & $192980$ \\
    \multicolumn{2}{|p{6cm}||}{Total number of stationary objects in a video} & $137485$ \\
    \multicolumn{2}{|p{6cm}||}{Total number of moving objects in a video} & $55495$ \\
    \hline
    \multicolumn{2}{|p{6cm}||}{Number of tracks in a video} & $[4,19]$ \\
    \multicolumn{2}{|p{6cm}||}{Temporal length of tracks (in frames)} & $[19,600]$ \\
    \hline
  \end{tabular}
  \vspace{-4mm}
\end{table}

\vspace{-2mm}
\section{Object detection}\label{sec:assume_static}

\begin{figure}
  \centering
  \includegraphics[width=0.85\linewidth]{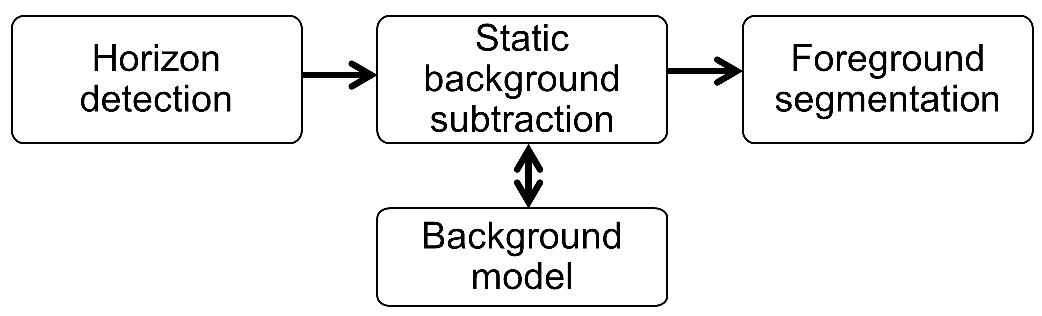}\vspace{-2mm}
  \caption{\vspace{-2mm}General pipeline of maritime EO data processing for object detection.}\label{fig:objectdetection}\vspace{-5mm}
\end{figure}

For object detection in maritime EO data processing, each frame of the EO video stream is considered independently without taking temporal information into account. The general framework of object detection approaches in maritime scenarios is shown in Fig. \ref{fig:objectdetection}. It consists of three main steps, viz., horizon detection, background subtraction, and foreground segmentation, discussed in the following subsections.

\vspace{-2mm}
\subsection{Horizon detection}\label{subsec:horizon}

There are three main approaches for horizon detection $-$ projection based, region based, and hybrid approach. 
Fig. \ref{fig:horizon_issues}(a) shows three examples of maritime images with horizon. Image 1 has two very low contrast targets close to a blurry horizon. Image 2 has horizon characterized by good contrast between the sky and water. Image 3 does not have a well defined horizon although the presence of the skyline may be a useful cue for its detection. However, the image suffers from false horizontal line created by the wakes of two targets, which is likely to be confused as horizon. Due to a variety of available cues as well as challenges, we use these images as examples to demonstrate the strengths and weaknesses of different horizon detection methods.

\subsubsection{Projections from edge map}

In these methods, first the edge map of the image is computed using edge detectors \cite{gonzalez2009digital}. It is then projected to another space where prominent line features in the edge map can be identified easily. Typically, the Hough and Radon transforms are used for such projections. Given the equation of a line:
\begin{equation}\label{eq:Hough}
  x \cos(\theta) + y \sin(\theta) = \rho
\end{equation}
Each edge pixel with coordinates $(x,y)$ is transformed into a curve in the Hough space $(\theta,\rho)$ using the projection \cite{gonzalez2009digital}:
\begin{equation}\label{eq:HoughProjection}
 \small H(\theta,\rho)\!\! =\!\! \int\!\!\!\!\int_{x,y} {\!\!\Big(\!1-\delta\big(\!I(x,y)\!\big)\!\Big)\delta(x \cos \theta + y \sin \theta -\rho) \,dx\,dy}
\end{equation}
where $\delta$ represents the Dirac delta function and $I(x,y)$ is the edge map. This is analogous to computing the 2D histograms of $(\theta,\rho)$. Cells in the $(\theta,\rho)$ histogram corresponding to few largest values of $H(\theta,\rho)$ determines the parameters of the line.

The transformation into Radon space is achieved by \cite{gonzalez2009digital}:
\begin{equation}\label{eq:RadonProjection}
  R(\theta,\rho) = \int\!\!\!\!\int_{x,y} {I(x,y)\delta(x \cos(\theta) + y \sin(\theta) -\rho) \,dx\,dy}
\end{equation}
Similar to the Hough space, cells in $(\theta,\rho)$ with the highest number of entries in $R(\theta,\rho)$ are the parameters of the line.

\begin{figure}
  \centering
  \includegraphics[width=\linewidth]{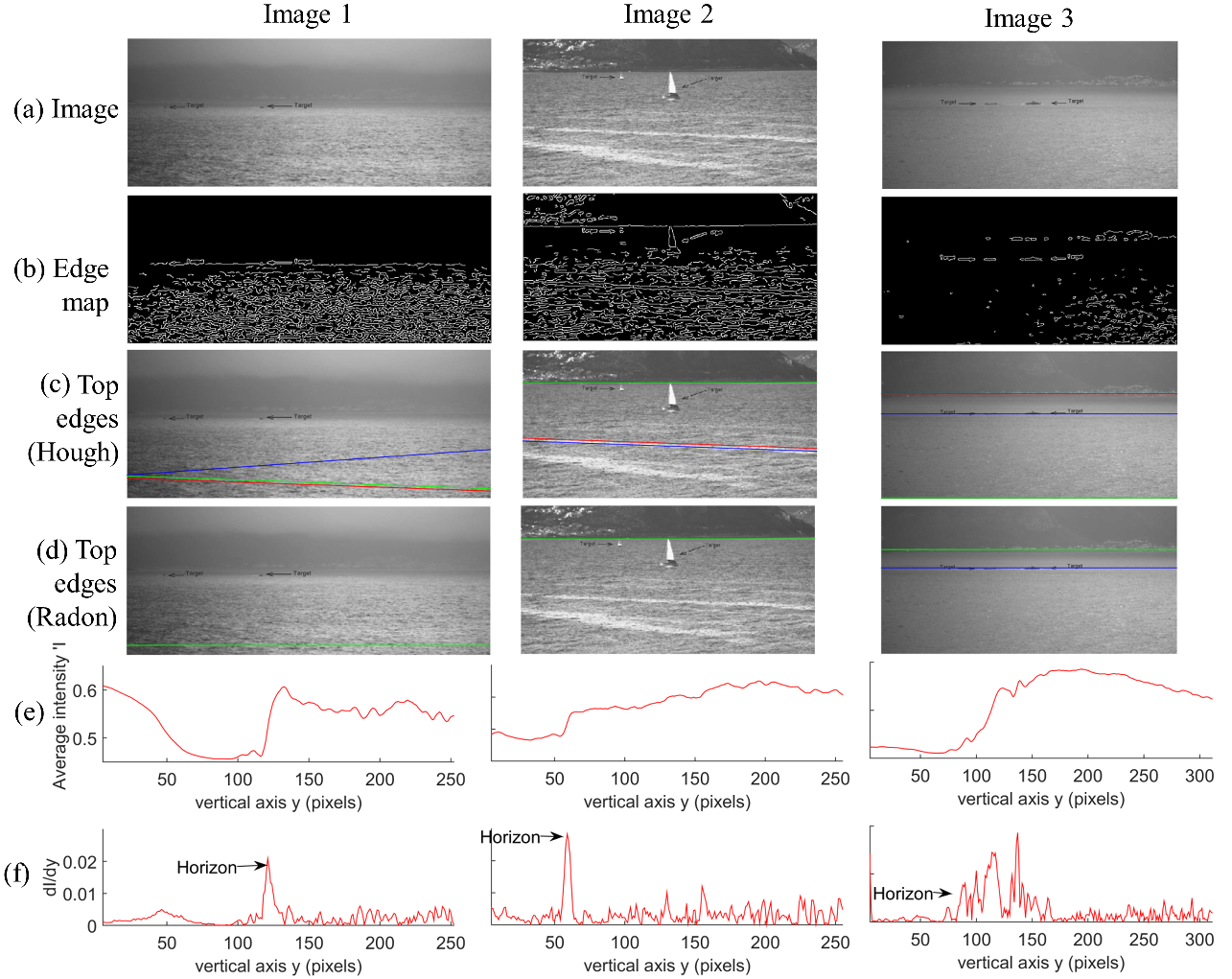}\vspace{-2mm}
  \caption{Three example images (row a) from \cite{szpak2011maritime} and their edge maps (row b) are used for studying the problem of horizon detection. The top 3 candidates, with largest strengths in Hough and Radon spaces are shown in rows (c,d) using colored lines. Average intensity profiles in the vertical direction are shown in (e). The gradients of intensity profiles in (e) are shown in (f).}\label{fig:horizon_issues}\vspace{-3mm}
\end{figure}

While the simplicity of these approaches makes them popular, projective transforms are sensitive to preprocessing such as histogram equalization and filtering before the extraction of the edge map ~\cite{gonzalez2009digital,bao2005vision}. Further, they can detect the horizon only if it appears as a prominent line feature in the edge map. Thus, as shown in Fig. \ref{fig:horizon_issues}(c,d), both Hough and Radon transforms perform poorly for image 1 but detect the horizon in image 2. Although the edge map of image 3 (Fig. \ref{fig:horizon_issues}(b)) does not have significant features corresponding to the horizon, the city skyline provides sufficient number of edge pixels parallel and close to the horizon, enabling a rough detection of horizon. Notably, the wake creates a dark horizontal stripe close to the targets in image 3 which causes detection of the line corresponding to the wakes as well.

\subsubsection{Region based horizon detection}

The intensity variations in the region of the horizon are higher compared to sky or sea regions alone. In Fig. \ref{fig:horizon_issues}(e,f), the mean intensity along the vertical axis and its gradient are plotted for images 1$-$3. The regions of horizon are characterized by significantly large intensity changes in each of the three images, although the intensity gradient itself is not sufficiently conclusive of the horizon in image 3. Such localized intensity characteristics are used for detecting horizon, especially in unmanned aerial vehicles \cite{demonceaux2006omnidirectional,ettinger2002towards,todorovic2004vision}. Quite often, the pixels in an image are classified as belonging to sky and sea (or ground) \cite{bhanu1990model}.

This is a three step procedure. The first step is to use a local smoothing operator, such as top-hat filter \cite{Zhou2014}, median filter \cite{gershikov2013horizon}, mean filter \cite{Wang2011FuzzyClutter}, Gaussian filter \cite{sheng1999real}, or standard deviation filter \cite{Wang2011FuzzyClutter}. The second step is to approximate these local statistics with sum of Gaussian functions or polynomial functions \cite{van2009polynomial,van2000detection}, where each function represents distribution of one region, such as the sea region or the sky region. More complex representations of the regions, such as linear discriminant analysis \cite{todorovic2004vision_ieee}, textures \cite{todorovic2004vision_ieee}, covariances \cite{gershikov2013horizon,ettinger2003vision}, and eigenvalues \cite{ettinger2003vision}, may be used. In the last step, the boundary of two classified regions is identified as horizon. We note that region based techniques inherently assume apriori information, such as suitable statistical representations or machine learning of the trend of intensities at the horizon.

Instead of the second and third steps, Bouma et. al \cite{bouma2008automatic} used high intensity gradient to conclude the common boundary of sea-sky regions and used it as horizon. A more robust version of this approach employed multi-scale approach \cite{romeny2003front}.\textcolor{white}{\ref{tab:horizon}}

\begin{table*}
\caption{Horizon detection approaches.}\label{tab:horizon}\vspace{-2mm}
  \centering
  \begin{tabular}{|l||p{2.4cm}|p{4.3cm}|p{6.5cm}|}
     \hline
     {} & Methods & Advantages & Disadvantages \\
     \hline
     \hline
     Projection from edge  & Radon transform, & {$\odot$ Simple } & {$\otimes$ Sensitive to preprocessing}\\
     map {\cite{bloisi2011automatic,wei2009automated,tang2013research}} & Hough transform & $\odot$ Mathematically well-defined & $\otimes$ Work for prominent well-defined linear horizon only\\
     {} & {} & {} & {$\otimes$ Horizon may not be the most prominent}\\
     \hline
     Region based & Median, correlation, & $\odot$ Work for blurred horizon as well & $\otimes$ Requires statistical apriori knowledge \\
     \cite{van2009polynomial,bhanu1990model,bouma2008automatic,Wang2011FuzzyClutter} & covariance & $\odot$ Suitable for IR images & $\otimes$ Based on statistics \\
     \hline
     Hybrid & & $\odot$ More accurate & $\otimes$ More complex\\
     \cite{fefilatyev2006horizon,bouma2008automatic,fefilatyev2012algorithms} & & $\odot$ More robust to low-contrast images & $\otimes$ More computation intensive\\
     \hline
   \end{tabular}\vspace{-2mm}
\end{table*}

\begin{figure}
  \centering
  \includegraphics[width=\linewidth]{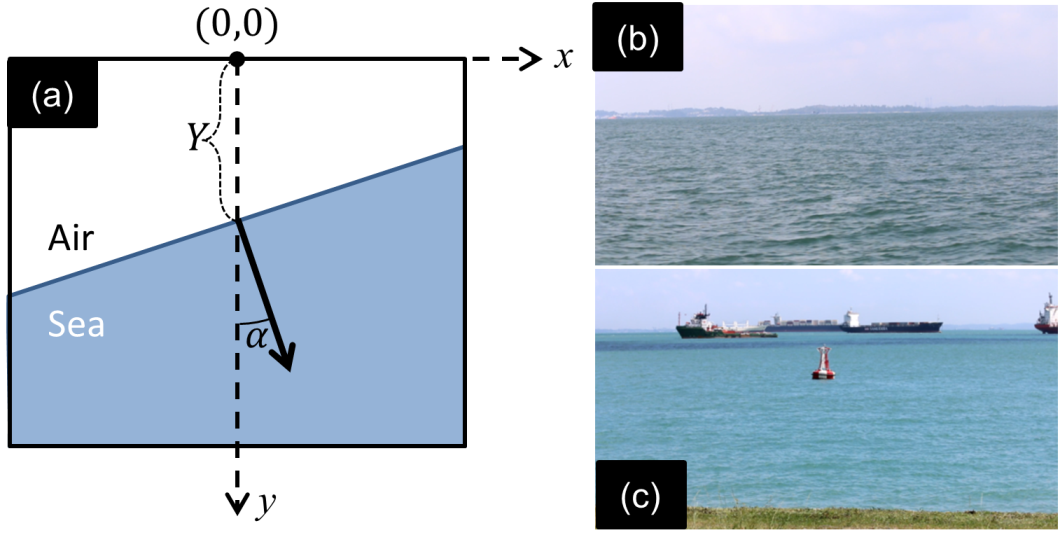}\vspace{-2mm}
  \caption{Representation of horizon for quantitative comparison of horizon detection approaches is shown in (a). Representative frames from on-board and on-shore videos are shown in (b,c) respectively.}\label{fig:hori_rep}\vspace{-5mm}
\end{figure}

\subsubsection{Hybrid methods}

The above methods are ineffective for Image 3 in Fig. \ref{fig:horizon_issues}. In such cases, hybrid methods are useful. In the hybrid approach of \cite{fefilatyev2012detection}, for each candidate generated by a projection-based method, the regions above and below the candidate line were considered as hypothetical sky and sea regions, and their statistical distributions were computed. The candidate that gives maximum value of the Mahalanobis distance between the distributions of the hypothetical sea and sky regions is chosen as horizon. In \cite{fefilatyev2006horizon}, local statistical features were used, but explicit representations of sea and sky regions were not employed. Then, using a set of training images and machine learning techniques, features representing horizon were learnt directly. Recent algorithms have combined multi-scale filtering and projection based approaches for providing state-of-the-art results \cite{prasad2016mscm,prasad2016muscowert}.

\subsubsection{Comparison of methods for horizon detection}

A qualitative comparison of the methods is provided in Table \ref{tab:horizon}. For quantitative comparison on Singapore Marine dataset, we use the representation of horizon as shown in Fig. \ref{fig:hori_rep}(a). $Y$ is the distance between the center of the horizon and the upper edge of the frame. $\alpha$ is the angle between the normal to the horizon and the vertical axis of the frame. The ranges and standard deviations of $Y$ and $\alpha$ are quite large for on-board videos (see Table \ref{tab:dataset}), making them significantly more challenging than on-shore videos. On-board videos are challenging due to presence of land features very close to horizon, while on-shore videos are also challenging because of the occlusion of horizon by vessels and the presence of wakes in the foreground. Frames from both types of videos are shown in Fig. \ref{fig:hori_rep}(b,c).

\begin{table}
  \centering
  \caption{Quantitative comparison of methods for horizon detection. The smallest error in each column is indicated in bold.} \label{tab:quant_hori}\vspace{-2mm}
  \begin{tabular}{|p{0.755cm}||p{0.445cm}|p{0.3cm}|p{0.3cm}|p{0.3cm}||c|p{0.3cm}|p{0.3cm}|p{0.3cm}||c|}
    \hline
    \multirow{3}{*}{}  & \multicolumn{4}{c||}{Position error} & \multicolumn{4}{c|}{Angular error}& \multirow{3}{0.75cm}{Time/ frame (s)}\\
    {} & \multicolumn{4}{c||}{$|Y_{GT}-Y_{est}|$} & \multicolumn{4}{c|}{$|\alpha_{GT}-\alpha_{est}|$} & {}\\
    \cline{2-9}
    {} &
    \hspace{-2mm}Mean & \hspace{-2mm}Q25 & \hspace{-2mm}Q50 & \hspace{-2mm}Q75 &
    \hspace{-2mm}Mean & \hspace{-2mm}Q25 & \hspace{-2mm}Q50 & \hspace{-2mm}Q75 & \\
    \hline
    \hline
    \multicolumn{10}{|c|}{On-board videos}\\
    \hline
    \hspace{-2mm}Hough &
    219 & 131 & 220 & 295 &
    2.6 & 0.6 & 1.7 & 3.4 & \textbf{0.3}\\
    \hspace{-2mm}Radon &
    372 & 213 & 362 & 517 &
    40.6 & 1.5 & 3.4 & 87.7 & {2.7}\\
    \hline
    \hspace{-2mm}MuSMF &
    269 & 156 & 283 & 379 &
    \textbf{1.8} & \textbf{0.5} & \textbf{1.2} & \textbf{2.5} & {0.9}\\
    \hspace{-2mm}ENIW &
    \textbf{120} & \textbf{63} & \textbf{116} & 166 &
    1.9 & \textbf{0.5} & \textbf{1.2} & \textbf{2.5}  & hours \\
    \hline
    \hspace{-2mm}FGSL &
    \textbf{120} & \textbf{63} & 117 & \textbf{165} &
    \textbf{1.8} & \textbf{0.5} & \textbf{1.2} & \textbf{2.5} & 12.8\\
    \hline
    \hline
    \multicolumn{10}{|c|}{On-shore videos}\\
    \hline
    \hspace{-2mm}Hough &
    208 & 26 & 194 & 354 &
    \textbf{1.2} & 0.2 & 0.7 & 1.5 & \textbf{0.2}\\
    \hspace{-2mm}Radon &
    313 & 28 & 359 & 549 &
    32.9 & 0.2 & \textbf{0.4} & 88.1 & {2.0}\\
    \hline
    \hspace{-2mm}MuSMF &
    \textbf{60} & 25 & \textbf{49} & \textbf{85} &
    \textbf{1.2} & 0.2 & \textbf{0.4} & \textbf{1.1} & 0.9\\
    \hspace{-2mm}ENIW &
    121 & 15 & 94 & 163 &
    \textbf{1.2} & 0.2 & \textbf{0.4} & 1.3 & hours \\
    \hline
    \hspace{-2mm}FGSL &
    112 & \textbf{12} & 91 & 162 &
    \textbf{1.2} & 0.2 & \textbf{0.4} & \textbf{1.1} & 12.3\\
    \hline
  \end{tabular}\vspace{-4mm}
\end{table}

We use position error $|Y_{GT} - Y_{est}|$ and angular error $|\alpha_{GT} - \alpha_{est}|$ as performance metrics for horizon detection. We provide comparison of Hough transform \cite{gershikov2013horizon} (referred to as Hough), Radon transform \cite{gonzalez2009digital} (Radon), mulit-scale median filter \cite{bouma2008automatic} (MuSMF), Ettinger et. al's method \cite{ettinger2002towards} (ENIW), and Fefilatyev et. al's method \cite{fefilatyev2012detection} (FGSL). Hough and Radon are projection-based, MuSMF and ENIW are region-based, and FGSL is a hybrid method. We have implemented MuSMF, ENIW, and FGSL since their codes are not available.

The comparison results are given in Table \ref{tab:quant_hori}. It is notable that the error in $Y$ is more severe for all the methods as compared to the error in $\alpha$. Projection based methods show poorest performance and statistical methods perform better. MuSMF performs the best for the on-shore videos , FGSL which uses both Hough transform and statistical distance measure for identifying the horizon performs the best for the on-board videos. MuSMF is parallelizable, with a possibility of making it about 10 times faster.

\vspace{-2mm}
\subsection{Static background subtraction}\label{subsec:static_background}

There is a large corpus of works related to background subtraction that originated from the computer vision community (see \cite{bouwmans2014traditional} for a review). Maritime background subtraction can be considered under two scenarios: open seas and close to port/harbor. In the former, the challenge arises due to the dynamic nature of the water background in the form of waves, wakes, and debris. In the latter, static structures such as buildings or stationary vessels pose a challenge. The current literature in maritime background subtraction almost exclusively deals with the case of open seas.

The challenge of dynamic background is largely alleviated if long-wave infrared sensor, such as forward looking infrared (FLIR), is used because it maps the temperature of water, which is relatively uniform despite the dynamicity of water. This suppression of dynamicity occurs to a smaller extent in near-infrared and mid-wave infrared wavelengths \cite{robert2007multispectral}. Thus, static background subtraction techniques have better performance in long-wave infrared regime than in the visible spectrum. In the following, we describe the various background subtraction techniques based on background models and their corresponding learning strategies.

\begin{table*}
  \caption{Summary of static background subtraction approaches used in object detection.} \label{tab:static_background}\vspace{-2mm}
  \centering
  \begin{tabular}{|p{2cm}||p{3.75cm}|p{3.5cm}|p{2.25cm}|p{4.0cm}|}
    \hline
    \multicolumn{1}{|c||}{Approach} & \multicolumn{1}{c|}{Model} & \multicolumn{1}{c|}{Learning} & \multicolumn{1}{c|}{Advantages} & \multicolumn{1}{c|}{Disadvantages} \\
    \hline
    \hline
    Single image statistics \cite{Ren2012,Zhou2014,bhanu1990model,bouma2008automatic,van2000detection,Chen2014,smith1999identification,Wang2011FuzzyClutter,Yao2013,ettinger2002towards} & Histogram correlation, polynomial functions fitting to strips parallel to horizon, spatial co-occurrence of intensities, spatial filtering & No temporal learning, only spatial patches/strips are used, initial background typically learnt as pixels using spatial standard deviations & Simple, no learning involved, no need of memory & Cannot deal with multi-modal approaches, does not use any form of temporal information\\
    \hline
    Gaussian mixture model (GMM) \cite{wang2014aquatic,fefilatyev2012detection,frost2013detection,zhang2012visual} & Probability of intensity at a background pixel is a combination of Gaussian functions & Supervised learning using background labelled images or videos, adaptive learning can be used to update GMM & Adapts for multi-modal background and illumination changes & Requires supervised learning using a suitable dataset, adaptive learning may be complicated\\
    \hline
    Bayes classifier \cite{socek2005hybrid} & \footnotesize{Compute the Bayes conditional probabilities of the pixel being background (or foreground) given an observed feature vector} & Supervised learning through a suitable training dataset & Classification is simple & Learning is complicated and sensitive to the training dataset\\
    \hline
    Feature based background classifier \cite{zhu2010novel} & Compute the feature attributes for every pixel and determine distance from pre-learnt class features & Supervised learning of class features using a suitable training dataset with both positive and negative samples & Robust due to multiple attributes class representation & Computation intensive learning, testing more complex than other methods above, multi-class may represent wakes, foam, clouds, etc.\\
    \hline
  \end{tabular}\vspace{-4mm}
\end{table*}

\subsubsection{Image statistics}

Methods in this category use statistical information in a single infrared image. One of the earlier methods is by Bhanu and Holben [15], which modeled an image in terms of gray scale intensity and edge magnitude with the aim of segmenting the image into foreground and background using a relaxation function that gives a low value for background and a high value for the foreground. Similar idea was employed in \cite{jabri2000detection,mason2001using}, where, instead of the relaxation function used in \cite{bhanu1990model}, confidence maps \cite{jabri2000detection} and chi-squared measure of similarity \cite{mason2001using} were used to segment the image into background and foreground regions.

Smith and Teal \cite{smith1999identification} compared the histogram of gray level intensities in a pixel's vicinity with a histogram of intensities of a reference background. If the histograms were similar, then the pixel was assigned to background. The reference background was obtained from the image itself by computing standard deviation over a $3\times3$ window at each pixel and assigning the pixels with less standard deviation as the background. The method fails if the reference background is computed incorrectly or if the background has wakes and debris such that their histograms may not correlate with the reference background histogram.

Van den Broek et. al \cite{van2000detection} used horizon to determine the sea and sky regions. It was assumed that intensities in the sea region may vary perpendicular but not parallel to the horizon. Thus, mean and standard deviations of the intensities along thin strips parallel to the horizon were computed and polynomial functions fit upon the mean and standard deviations of the strips. Similar polynomial fitting was performed for the sky as well. These polynomial models for sea and sky were then used to compute a background map and subtract it from the image. This approach removed only low spatial frequency component from the image. Although a more robust approach proposed in \cite{meer2000robust} was used by \cite{bouma2008automatic} for visible range images, natural high spatial frequency components such as due to waves, sun, and clouds were still retained. Gal \cite{Gal2011} used a co-occurrence matrix approach to learn sea and sky patterns which were subtracted from the original image. Fefilatyev et. al \cite{fefilatyev2010tracking} used Gaussian low pass filtering in a narrow strip below horizon, followed by a color gradient filter to obtain regions of high color variations and finally applied a threshold computed using Otsu's method \cite{otsu1975threshold} to obtain the background.

Chen at. al \cite{Chen2014} considered suppressing repeated spatial patterns by suppressing peaks in the Fourier transform of the image. Spatio-spectral residue and phase map of Fourier transform of eigenvectors representing 80\% of input image were used in \cite{Ren2012} for background suppression. Multi-scale approaches, combined with low-pass filter extracting low spatial frequencies \cite{Hou2007,Guo2008}, which are representations of background, have also been found useful. For example, top-hat convolution filter, which is a low-pass filter, used in a multi-scale approach was shown to be effective in wake suppression \cite{Zhou2014}. Multi-scale spatio-spectral residue was used in \cite{Yao2013}. Wang and Zhang \cite{Wang2011FuzzyClutter} used multilevel filter and recursive Otsu approach \cite{otsu1975threshold} to detect and segment very small and either dark or bright targets from images with complex background.

\subsubsection{Gaussian mixture model (GMM)}\label{subsubsec:GMM}

Although wakes and foam appear distinct in the visible range images, they are not entirely suppressed even in infrared images. Thus, the histograms of both visible and infrared images are invariably multimodal. Gaussian mixture models (GMM) \cite{stauffer1999adaptive,wren1997pfinder} are suitable for representing multimodal backgrounds \cite{bloisi2009argos,wang2014aquatic,fefilatyev2012detection,frost2013detection,gupta2009adaptive}.
If a pixel belongs to the background, the probability $P(I)$ of observing an intensity $I$ at that pixel is given as:
\begin{equation}\label{eq:GMM}
  P(I)=\sum_i{w_i G(\mu_i,\sigma_{i})}
\end{equation}
where $G(\mu_i,\sigma_{i})$ represents $i^{\rm th}$ Gaussian distribution with mean $\mu_i$ and standard deviation $\sigma_{i}$, and $w_i$ is the weight of the $i^{\rm th}$ Gaussian distribution. Fefilatyev et. al \cite{fefilatyev2012detection} represented sky and sea regions as two Gaussian distributions (each being trivariate due to the red, green, and blue color channels) fitted with maximum possible separation between their means. In \cite{wang2014aquatic}, if the distance of the test pixel's intensity from the mean of the closest Gaussian distribution was within $2.5$ times its standard deviation, then the pixel was classified as background.

\subsubsection{Bayes classifier}

Socek et. al \cite{socek2005hybrid} used Bayes classifier approach of \cite{li2003foreground} for background estimation and suppression.
Given the feature vector at a test pixel $\bar v(p)$, if $P\big(p \in \bf B|\bar v(p)\big) > P\big(p \in \bf F|\bar v(p)\big)$, where $\bf B$ and $\bf F$ indicate the background and the foreground, then the test pixel was classified as the background. The likelihoods  $P\big(\bar v(p)|p \in \bf B \big)$ and $P\big(\bar v(p)|p \in \bf F \big)$ were determined through the histogram of the feature vector, learnt apriori. The supervised learning strategy enforced that a certain percentage of pixels should be classified as background using another background estimation method. It was found that the segmented frames contained too many noise-related and scattered pixels, which may be separately classified as outliers and removed, however at the expense of missing small objects that are few pixels wide.

\subsubsection{Feature based background classifier}

In \cite{zhu2010novel}, both foreground and background objects were considered as belonging to different known classes, viz., clouds, islands, coastlines, oceanic waves, and ships. It used a total of seven types of features, viz., shape compactness, shape convexity, shape rectangularity or eccentricity, shape moment invariants, wavelet-based features, multiple Gaussian difference features, and local multiple patterns (discussed more in section \ref{subsubsec:LBP}). It also considered what combinations of features were suitable for improving the detection accuracy of the different subclasses.

\begin{table}
  \centering
  \caption{Quantitative comparison of background subtraction approaches for on-shore videos of Singapore Marine dataset. The best values in each column are indicated in bold.}\label{tab:quant_bck}
  \begin{tabular}{|p{0.95cm}||p{0.41cm}|p{0.35cm}|p{0.35cm}|p{0.41cm}||p{0.41cm}|p{0.35cm}|p{0.41cm}|p{0.41cm}||p{0.4cm}|}
     \hline
     \multirow{3}{*}{} & \multicolumn{4}{c||}{Precision ($\times 10^{-2}$)} & \multicolumn{4}{c||}{Recall ($\times 10^{-2}$)} & \multirow{3}{0.4cm}{\scriptsize{Time/ frame (ms)}}\\
     \cline{2-9}
     {} & \multirow{2}{*}{\hspace{-1mm}Mean} & \multirow{2}{*}{Q25} & \multirow{2}{*}{Q50} & \multirow{2}{*}{Q75} & \multirow{2}{*}{\hspace{-1mm}Mean} & \multirow{2}{*}{Q25} & \multirow{2}{*}{Q50} & \multirow{2}{*}{Q75} & \\
     {} & {} & {} & {} & {} & {} & {} & {} & {} & {}\\
     \hline
     \hline
     \multicolumn{10}{|c|}{Static background subtraction}\\
     \hline
     \scriptsize{\hspace{-2mm}HistComp} &
     {$\,\,$0.01} & {0.00} & {0.00} & {$\,\,$0.01} &
     {$\,\,$0.06} & {0.00} & {$\,\,$0.02} & {$\,\,$0.05} & $\sim$s\\
     \scriptsize{\hspace{-2mm}StatGMM} &
     {$\,\,$0.01} & {0.00} & {0.00} & {$\,\,$0.01} &
     {$\,\,$0.08} & {0.00} & {$\,\,$0.00} & {$\,\,$0.03} & $\sim$s\\
     \hline
     \hline
     \multicolumn{10}{|c|}{Dynamic background subtraction}\\
     \hline
     \scriptsize{\hspace{-2mm}TempMean} &
     {$\,\,$0.07} & {0.00} & {0.00} & {$\,\,$0.00} &
     {$\,\,$0.03} & {0.00} & {$\,\,$0.00} & {$\,\,$0.00} & $\,\,\,$43\\
     \scriptsize{\hspace{-2mm}AdaMed} &
     {$\,\,$0.25} & {0.06} & {0.13} & {$\,\,$0.26} &
     \textbf{14.72} & \textbf{6.96} & \textbf{11.00} & \textbf{19.87} & $\,\,\,$61\\
     \scriptsize{\hspace{-2mm}GMM} &
     {$\,\,$0.56} & {0.05} & {0.29} & {$\,\,$0.67} &
     {$\,\,$9.71} & {1.91} & {$\,\,$7.96} & {13.77} & $\,\,\,$54\\
     \scriptsize{\hspace{-2mm}KDE} &
     {$\,\,$0.59} & {0.08} & {0.48} & {$\,\,$0.89} &
     {$\,\,$8.93} & {1.87} & {$\,\,$9.52} & {13.31} & $\,\,\,$83 \\
     \scriptsize{\hspace{-2mm}OptFlow} &
     \textbf{11.64} & \textbf{1.65} & \textbf{7.21} & \textbf{14.50} &
     {13.35} & {0.91} & $\,\,${6.85} & {17.99} & 360 \\
     \scriptsize{\hspace{-2mm}IMBS} &
     {$\,\,$0.86} & {0.33} & {0.62} & {$\,\,$0.99} &
     {$\,\,$7.67} & {2.23} & {$\,\,$6.62} & {11.15} & 156 \\
    \hline
    \hline
     \multicolumn{10}{|c|}{CV methods for background subtraction}\\
     \hline
     \scriptsize{\hspace{-2mm}LBP} &
     {$\,\,$0.31} & {0.04} & {0.27} & {$\,\,$0.45} &
     {$\,\,$3.79} & {0.67} & {$\,\,$1.88} & {$\,\,$4.49} & 629\\
     \scriptsize{\hspace{-2mm}LBSP} &
     $\,\,$\textbf{8.00} & {0.04} & \textbf{3.36} & $\,\,$\textbf{9.29} &
     {$\,\,$6.08} & {0.03} & {$\,\,$2.73} & {$\,\,$8.92} & 589\\
     \scriptsize{\hspace{-2mm}\scriptsize{FuzzGMM}} &
     {$\,\,$0.01} & {0.00} & {0.00} & {$\,\,$0.01} &
     {$\,\,$0.06} & {0.00} & {$\,\,$0.02} & {$\,\,$0.05} & $\,\,\,$63\\
     \scriptsize{\hspace{-2mm}FAdaSOM} &
     {$\,\,$0.55} & \textbf{0.18} & {0.32} & {$\,\,$0.85} &
     {11.01} & {3.07} & {$\,\,$9.67} & {13.89} & 133\\
     \scriptsize{\hspace{-2mm}EigHMM} &
     {$\,\,$0.26} & {0.05} & {0.14} & $\,\,${0.30} &
     \textbf{14.55} & \textbf{4.68} & \textbf{12.63} & \textbf{18.48} & 245 \\
     \hline
   \end{tabular}\vspace{-4mm}
\end{table}

\vspace{-2mm}
\subsection{Foreground segmentation}\label{subsec:seg_before_track}

In traditional maritime data processing, applying morphological operations such as identifying closed boundaries after background subtraction were considered sufficient for foreground segmentation \cite{socek2005hybrid,sumimoto1994machine,westall2008evaluation} and the segmented contours were used as detected objects. Useful morphological operations for maritime human rescue problem have been adapted in \cite{westall2008evaluation} from \cite{casasent1997detection,deshpande1999max}.
Several interesting edge based morphological segmentation techniques for detecting objects have been discussed in \cite{bhanu1990model}. All are based on considering object segmentation as a two-class gray level problem in which objects belong to one set of gray levels and the background to the other. A further constraint is that the gradient inside an instance of each class is close to zero and gradients are high only along the edges. The underlying assumption in all morphological segmentation approaches is that the objects are not be occluded and are separate enough such that their boundaries may not merge.

\vspace{-2mm}
\subsection{Comparison of static background subtraction techniques}\label{sec:quant_static_bkgnd}

A qualitative comparison is given in Table \ref{tab:static_background}. Here, we present quantitative comparison of a few static background subtraction techniques. The foreground is morphologically obtained after static background subtraction and enclosed in bounding boxes. They are compared against the bounding boxes of the objects annotated as ground truth. The performance is evaluated using intersection over union (IOU) ratio of the bounding boxes, defined as
\begin{equation}\label{eq:IOU}
  {\rm IOU}\left({O^{\rm GT}_i,O^{\rm det}_j}\right)=\frac{{\rm Area}\left({O^{\rm GT}_i \cap O^{\rm det}_j}\right)}{{\rm Area}\left({O^{\rm GT}_i \cup O^{\rm det}_j}\right)}
\end{equation}
where $O^{\rm GT}_i$ and $O^{\rm det}_j$ are the bounding boxes of $i$th ground truth (GT) object and the $j$th detected object and $\rm Area$ denotes the number of pixels. If more than one detected objects overlap with a GT object, the detected object with maximum overlap with the GT object is considered associated with the GT and dropped from further associations. The unassociated objects or associated objects with IOU less than 0.5 (based on \cite{pascal-voc-2012}) are labelled as false positives (FPs). The remaining associated objects are true positives (TPs). The GT objects that are not associated to the any detected objects are labelled false negatives (FNs). $N_{\rm TP}$ is the number of TPs in the video, analogously for $N_{\rm FP}$ and $N_{\rm FN}$. Precision and recall are computed as
\begin{eqnarray}
  {\rm Precision} &=& {N_{\rm TP}}/{(N_{\rm TP}+N_{\rm FP})} \\
  {\rm Recall} &=& {N_{\rm TP}}/{(N_{\rm TP}+N_{\rm FN})}
\end{eqnarray}

Unfortunately, codes for static background subtraction in maritime research are not available. We implemented histogram comparison method (HistComp) of Smith and Teal \cite{smith1999identification} since sufficient implementation details were available. Further, we implemented a GMM for background subtraction in an image (StatGMM). We used the reference background computed in \cite{smith1999identification} for fitting one GMM each for the red and green color channels. Blue channel was not considered since the histogram of the blue channel's data is very narrow compared to the histograms of the other channels for maritime images. Kullback Leibler (KL) divergence \cite{kullback1951information} of these histograms from the GMMs are determined. In the KL map of each pixel, positive values indicate that the local histograms of the red and green color values are close to the static GMMs. This implementation is intended to serve as the worst performance scenario of GMM for maritime on-shore videos.

The comparison results are presented in Table \ref{tab:quant_bck}. The histogram comparison technique of \cite{smith1999identification} and static GMM perform almost similar, neither providing adequate precision and recall. We note that HistComp was tested on maritime intensity images acquired using low-resolution infrared cameras, and the high definition visible range color videos in Singapore-Marine dataset may have caused the poor performance. Similarly, most static background subtraction techniques were tested on low-resolution intensity images. Thus, these methods may not be suitable for high-resolution maritime imaging.

\begin{figure}
  \centering
  \includegraphics[width=0.9\linewidth]{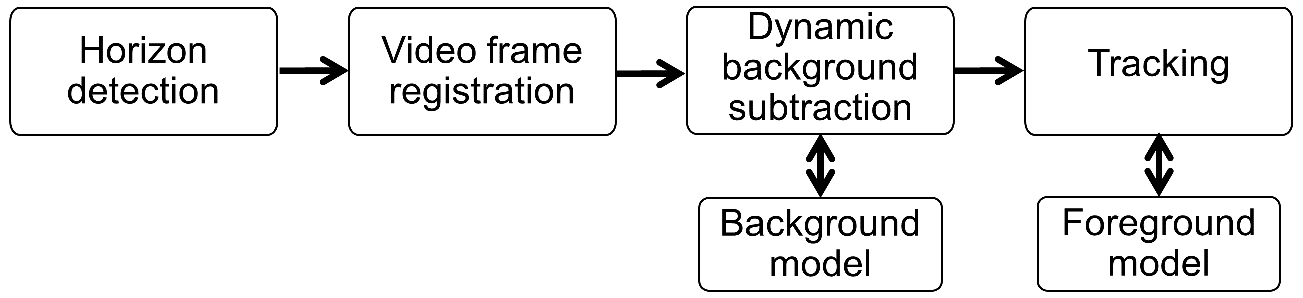}\\
  \caption{General pipeline of maritime EO data processing for object tracking.}\label{fig:objecttracking}\vspace{-5mm}
\end{figure}

\vspace{-2mm}
\section{Object tracking}\label{sec:assume_dynamic}

In much of the literature related to object tracking in maritime environment, the problem of object tracking is reduced to the problem of object detection in every frame. We differentiate between object detection algorithms and object tracking algorithms in that the latter use (i) temporal information across frames, e.g. optical flow, and (ii) employ dynamic background subtraction algorithms for more robust modeling of the background. A typical pipeline for maritime object tracking is shown in Fig. \ref{fig:objecttracking}. Below, we discuss each of the modules in the pipeline.

\begin{figure}
  \centering
  \includegraphics[width=0.95\linewidth]{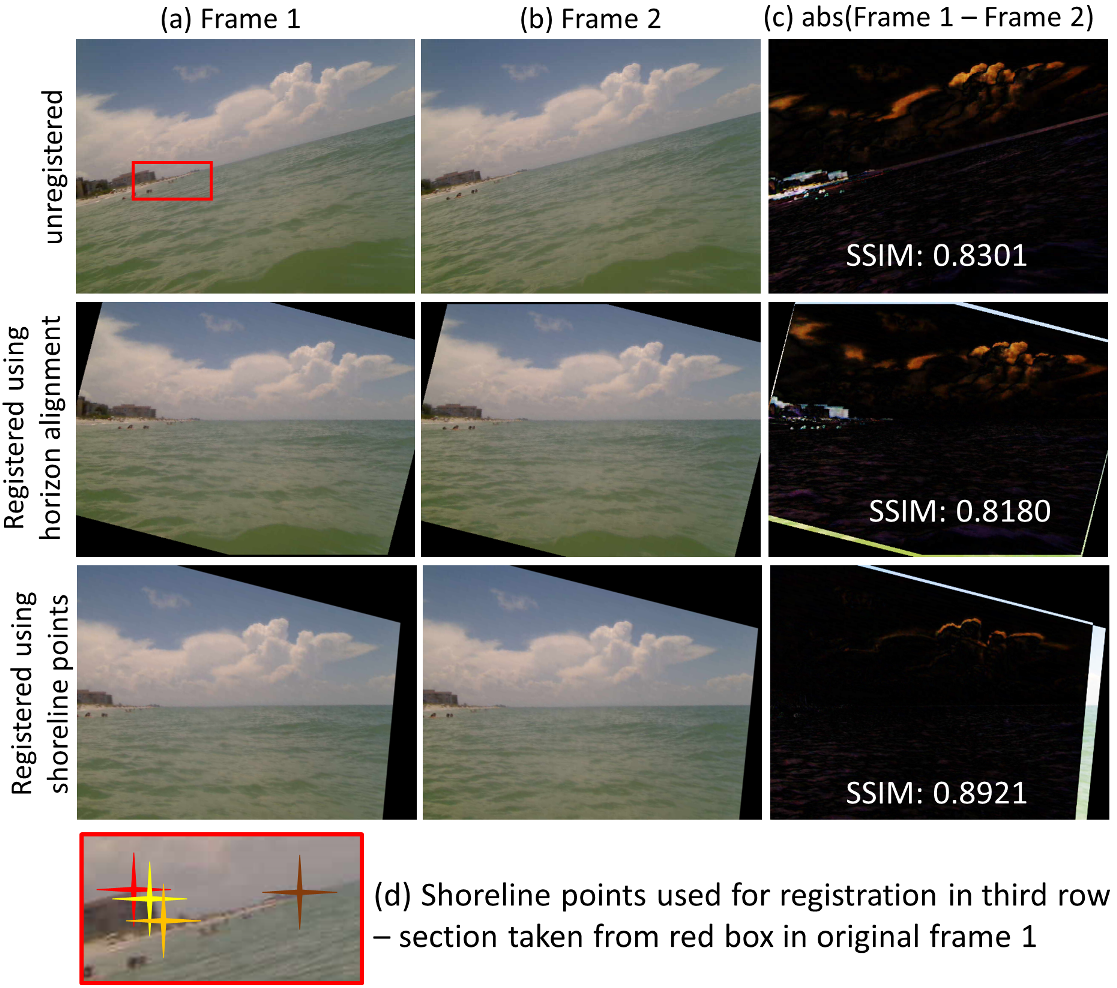}\vspace{-2mm}
  \caption{The top row shows two consecutive frames and their difference. Second row: result of registration results horizon. Third row: registration results using just four fixed points on the shoreline. Fourth row: The four points used for registration in the third row. Saturation and brightness of the difference images in column (c) have been enhanced for better illustration. SSIM \cite{wang2004ssim} for the image pairs is provided in column (c).}\label{fig:registrationprobelm2}\vspace{-2mm}
\end{figure}

\vspace{-2mm}
\subsection{Utility of horizon detection}

We discuss the use of horizon detection in object tracking. In object tracking, the main purpose of horizon detection is to allow for registration over consecutive frames and compensate for the motion of camera or its mounting base (such as due to turbulence of water inducing motion in a boat). Horizon may also be used for determining special conditions for detecting objects close to horizon. For example, \cite{voles2000nautical} used smaller video bricks close to horizon as compared to elsewhere. In some cases, horizon was used as an indicator of the distance between the camera and the vessel being tracked \cite{van2008discriminating} or for motion segmentation of the vessel \cite{brox2006variational}. However, sensitivity of the distance computation to the error in horizon detection was noted as severely restrictive in \cite{van2008discriminating,brox2006variational}.

\vspace{-2mm}
\subsection{Registration}\label{subsec:register}

\begin{figure}
  \centering
  \includegraphics[width=\linewidth]{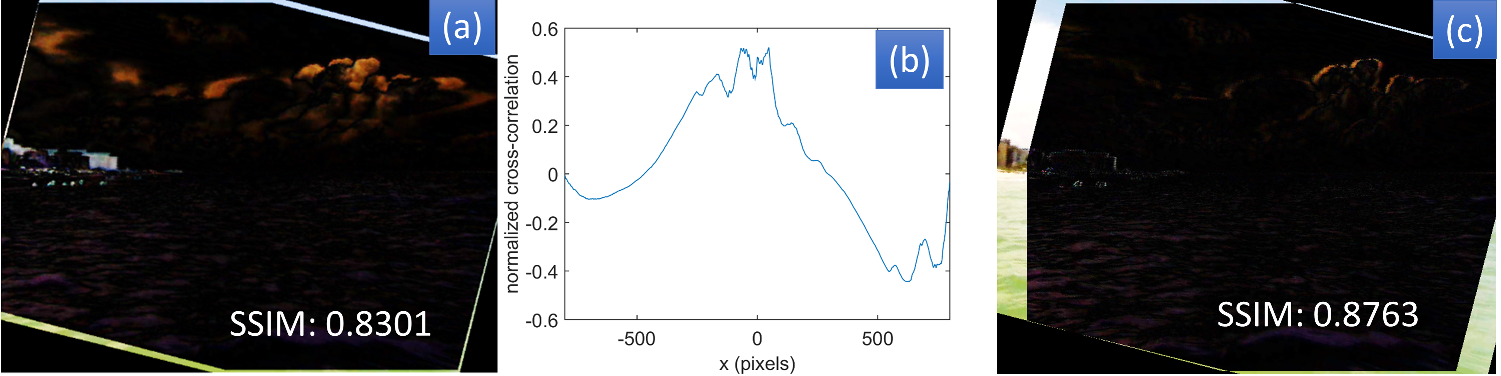}\vspace{-2mm}
  \caption{Registration using cross-correlation of strip around the horizon. (a) The difference image obtained by registration using horizon only, reproduced from Fig. \ref{fig:registrationprobelm2}. (b) The cross-correlation function of Fefilatyev \cite{fefilatyev2012algorithms}. (c) The difference image after horizontal shift of 48 pixels, identified as the peak in (b). Saturation and brightness of the difference images (a,c) have been enhanced for better illustration. SSIM \cite{wang2004ssim} for the image pairs is provided.}\label{fig:Fefilatyev}\vspace{-3mm}
\end{figure}

In maritime scenario, consecutive frames may experience large angular or positional shift. The angular difference may be due to yaw, roll, and pitch of the vessel. The positional shift comes from the fact that the sensor itself is not necessarily mounted at the effective center of motion of the vehicle. If the horizon is present, the discrepancy in roll and pitch can be corrected by the change of angle and position of the horizon, respectively. However, {the correction of the yaw cannot be achieved.} To illustrate this point, consider two consecutive frames of a video shown in Fig. \ref{fig:registrationprobelm2}. Horizon based registration is partially effective (2nd row, 3rd image) but the mismatch along the horizontal direction indicates that the shift in yaw is not corrected. In order to correct for the yaw, we can use additional features from the scene that can help in registration as shown in the third row of Fig. \ref{fig:registrationprobelm2}. However, such features are not available in scenes of high seas. Fefilatyev et. al \cite{fefilatyev2010tracking,fefilatyev2012algorithms} computed normalized cross-correlation in the horizontal direction along a narrow horizontal strip around the horizon in the images to be registered. The peaks of the normalized cross-correlation function indicated the amount of shift between the two frames. An example is given in Fig. \ref{fig:Fefilatyev}.

We present a quantitative comparison of registration techniques discussed above on the on-board videos in Singapore Marine dataset. We used the ground truth of horizon for performing horizon based registration. For registration using correlation technique of \cite{fefilatyev2012algorithms}, we found that strip of width 100 pixels centered at horizon gave the best result. Lastly, we used speeded up robust features (SURF) for registration using feature matching. We used all the features that could be matched between a pair of images to perform registration.

\begin{table}
\caption{Quantitative evaluation of registration approaches on on-board videos of Singapore Marine Dataset.}\label{tab:Quant_reg}\vspace{-2mm}
  \centering
    \begin{tabular}{|p{2.0cm}|p{0.5cm}|p{0.45cm}|p{0.45cm}|p{0.45cm}|p{0.45cm}||c|}
      \hline
      {} & Mean & Q25 & Q50 & Q75 & Q90 & \multirow{2}{1.15cm}{Time/frame (ms)}\\
      {} & { } & {} & {} & {} & {} & \\
      \hline
      \hline
      Unregistered &
      0.75 & 0.66 & 0.74 & 0.83 & 0.98 & $\,\,-\,\,$\\
      \hline
      \hline
      {Using horizon} &
      0.81 & 0.75 & 0.81 & 0.88 & 0.94 & $\,\,\,$68\\
      \hline
      {Correlation \cite{fefilatyev2012algorithms}} &
      0.81 & 0.75 & 0.81 & 0.89 & 0.94 & 465\\
      \hline
      {Feature matching} &
      0.75 & 0.66 & 0.73 & 0.83 & 1.00 & 279\\
      \hline
    \end{tabular}\vspace{-4mm}
\end{table}

We use structural similarity index metric (SSIM) \cite{wang2004image} to compute the similarity between two consecutive frames. It is specifically suitable for texture matching, and thus a good metric for maritime images with dynamic background. The mean SSIM for all the consecutive frame pairs is 0.75, as seen in Table \ref{tab:Quant_reg}. Even the Q25 value of SSIM is 0.66. Registration using horizon and using the cross-correlation technique \cite{fefilatyev2012algorithms} improves SSIM by 6-7\%. However, the Q90 values show a decrease, indicating that some consecutive frames are less similar after registration using horizon only or cross-correlation technique of Fefilatyev et. al \cite{fefilatyev2012algorithms}. On the other hand, registration using feature matching hardly improved SSIM, with improvement appearing in only a few frames as noted from the Q90 value of SSIM. \textit{The results indicate that the cross-correlation technique \cite{fefilatyev2012algorithms}, although simple, does not provide a significant advantage over registration using horizon for the current dataset although it was found to be effective for the videos taken from buoy mounted camera in \cite{fefilatyev2012algorithms}. We expect that this may be related to either the variety of structures seen along the horizon or the angle of the camera with the sea surface. On the other hand, registration using SURF features is not effective because of the lack of reliable stationary features in on-board maritime videos.}

\vspace{-2mm}
\subsection{Dynamic background subtraction}

As discussed in section \ref{subsec:static_background}, long-wave infrared sensors suppress dynamicity of water, which is amiss in videos acquired from visible range sensors. If static background methods are used for such videos, the dynamicity of water causes incorrect detections. Methods that explicitly model the background as being dynamic are more effective in this case. Here, we discuss the dynamic background subtraction approaches used for maritime EO videos processing.

\subsubsection{Relatively stationary pixels}\label{subsubsec:relative}

If a pixel corresponds to sky or water which is relatively stationary in the past few frames, the temporal distributions of intensities at a pixel over these frames are expected to be unimodal. The mean or median of the distribution is used to determine if the pixel belongs to the background or foreground \cite{hu2011robust,smith1999identification,Sang1998,Barnett1989,voles1999target}.
%
%
A simple threshold approach was used for learning the background across the frames in \cite{van2009polynomial}. At each pixel, $\rm L_p$ norm of intensities over a small temporal window $I_{\rm thresh}(x,t)=\| \big( I(x,t') - \tilde I(x,t')\big); \forall 0 \le t-t' \le T \|_{{\rm p}}$ was computed. Here, $x$ and $t$ represent the pixel and the current frame, $T$ is the size of temporal window, $I$ and $\tilde I$ represent the actual and fitted intensities respectively, and $\|\;\;\|_{\rm p}$ represents the $\rm L_p$ norm. The fitted intensity $\tilde I$ was obtained by fitting a polynomial over the measured intensities $I$ in the temporal window of size $T$. If $\big( I(x,t') - \tilde I(x,t')\big)<I_{\rm thresh}(x,t)$, the pixels were assigned to the background.

\begin{table*}
  \caption{Summary of dynamic background subtraction approaches for object tracking.}\label{tab:dynamic_background}\vspace{-2mm}
  \centering
  \begin{tabular}{|p{2.3cm}||p{3.25cm}|p{3.35cm}|p{3.5cm}|p{3.5cm}|}
    \hline
    \multicolumn{1}{|c||}{Approach} & \multicolumn{1}{c|}{Model} & \multicolumn{1}{c|}{Learning} & \multicolumn{1}{c|}{Advantages} & \multicolumn{1}{c|}{Disadvantages} \\
    \hline
    \hline
    {\scriptsize{Relatively stationary pixels \cite{van2009polynomial,hu2011robust,smith1999identification}, \cite{Sang1998}$-$\cite{voles1999target}}} & {Temporal filter extracts statistical representative of background} & {Sliding temporal window, change from reference, spatial smoothing} & {Simple, computation efficient, online learning} & {Cannot deal with highly dynamic backgrounds, e.g., wakes} \\
    \hline
    {Spatio-temporal filtering approaches \cite{strickland1997wavelet,tu2014infrared,Wang2006SVMWavelets}} & {Background is modelled as low spatial frequency component, albeit with temporal variation} & {Sliding temporal window and fixed spatial window (blobs or neighborhood) are used for filter parameters' update } & {Simple, online learning, computation efficient, robust to small dynamics and illumination variation} & {Cannot deal with highly dynamic backgrounds, e.g., wakes} \\
    \hline
    {GMM \cite{bloisi2009argos,gupta2009adaptive}} & {Intensities at a pixel as mixture of Gaussian distributions} & {Fitting GMMs on histograms of intensities over past few frames} & {Online, less memory intensive, simple, adaptive} & {Cannot accommodate complex intensity distributions and sudden illumination changes}\\
    \hline
    {Kernel density estimation}\cite{mittal2004motion} & {Background modelled as sum of kernels of adaptive spreads} & { Learnt through fitting over last few frames} & {Asymmetric kernels may be used, online learning, fast and adaptive, can deal with small illumination variations, wakes, and foam better than GMM} & {Kernels should be good representative or else several kernels may be needed, adaptive nature makes it sensitive to variations} \\ \hline
    {Optical flow} \cite{ablavsky2003background} & {Segment initial background, compare with background predicted by motion map} & {Learnt by spatial gradients' patterns over time as velocities} & {Suitable for wakes, big waves, and clouds} & {Not suitable for random wave motion, computation intensive} \\
    \hline
    {Multi-step approaches \cite{bloisi2014background,socek2005hybrid}}& \multicolumn{2}{l|}{Combination of more than one technique} & {More robust and versatile, often made adaptive and capable of dealing with wakes} & {Complicated, computation intensive, slow due to frequent feedback \textemdash  feed-forward steps} \\
    \hline
  \end{tabular}\vspace{-4mm}
\end{table*}

\subsubsection{Spatio-temporal filtering approaches}

Wavelet transformation was used in \cite{strickland1997wavelet,tu2014infrared} for suppressing the background. In \cite{Wang2006SVMWavelets}, wavelet transform and support vector machine on low frequency wavelets were used to detect objects, followed by correlation over 5 frames, and adaptive segmentation. Low frequency wavelets were assumed to contain less information of clutter and then the uncluttered background would not correlate over the frames, thus both clutter and background could be taken care of.

\subsubsection{Gaussian mixture models}
Gaussian mixture models have already been introduced in the context of stationary background in section \ref{subsubsec:GMM}. In Bloisi and Iocchi \cite{bloisi2009argos} fitted a trivariate GMM for RGB values at each pixel over last few frames. The pixel was labeled as foreground if its RGB values differed from the GMM by a threshold $t$ that depended upon the illumination conditions. 
Gupta et. al \cite{gupta2009adaptive} also learnt the Gaussian model over past few frames only, although using time-weighted intensity values. The time weighting allowed for the GMM to adapt to the changing conditions to a small extent. A test pixel was classified as background if the significance score, proportional to the square of distance between the intensity of the test pixel and the mean of the Gaussian model, was small.

\subsubsection{Kernel density estimation (KDE)}\label{subsubsec:KDE}

Mittal et. al \cite{mittal2004motion} considered the ocean as dynamic background which was subtracted to detect people on shore. It used kernel density estimation (KDE) for background subtraction, in which the background model is represented as:\vspace{-2mm}
\begin{equation}\label{eq:KDE}
  P(I_t)=\frac{1}{n}\sum_{i=1}^{n}{K(I_t,\kappa_i)}
\end{equation}
where $P(I_t)$ is the probability of the intensity at a time $t$ at a given pixel is $I_t$, $K(I,\kappa)$ is the kernel function for the intensity $I$ with kernel parameters specified by $\kappa$, and $n$ is the number of kernels. Typically, single parameter kernels are used. KDE is different from GMM in two respects. First, GMM uses Gaussian kernel whereas KDE allows the kernel to be asymmetric or have suitable statistical properties. Second, unlike GMM, KDE need not use supervised learning. Typically only a few frames are used for computing the probability distribution and the assumed kernels are fit upon it.

\subsubsection{Optical flow}
Optical flow methods learn the patterns of motion from the videos. The flow vectors computed by comparing adjacent frames are used to warp each frame with respect to a reference frame such that stationary components can be identified as background. Ablavsky et. al \cite{ablavsky2003background} used optical flow technique for maritime images, specifically to model wakes as background. From a given image frame, pre-learnt motion maps were used to predict next image frame and correlate it with the actual new frame. The pixels with high correlation were labelled as background. We note that optical flow is used as one component in their multi-module interconnected framework for background subtraction. The other components are Bayesian probabilistic background estimation, motion map filter, and coherence analyzer.

\subsubsection{Multi-step approaches}

These approaches combine more than one technique to achieve better background subtraction.

Independent Multimodal Background Subtraction (IMBS) was proposed by Bloisi et. al \cite{bloisi2014background}. It has three components. The first component is an on-line clustering algorithm. The RGB values observed at a pixel is represented by histograms of variable bin size. This allows for modelling of non-Gaussian and irregular intensity patterns. The second component is a region-level understanding of the background for updating the background model. The regions with persistent foreground for a certain number of frames are included as background in the updated background model. The third component is a noise removal module that helps in filtering out false detections due to shadows \cite{cucchiara2003detecting}, reflections \cite{rankin2004daytime}, and boat wakes. It models wakes as outliers of the foreground, forming a second background model specifically for such outliers.

Socek et. al \cite{socek2005hybrid} used a four-step process for background extraction: change detection, change classification, foreground segmentation, and background model learning and maintenance. Change detection was done by subtracting the incoming image from a reference (pre-learnt) stationary image. The detected change image was then analyzed to find if the changes correlate to the prediction of Bayesian background model \cite{li2003foreground} or are they likely to be foreground. The regions classified as foreground were then used with color-based segmentation approach to further strengthen the foreground estimation and thus contribute to more robust background detection. The background thus determined was used to update the reference stationary image and the Bayesian background model.

\subsubsection{Comparison of techniques for dynamic background subtraction}

A qualitative comparison of the dynamic background segmentation methods is given in Table \ref{tab:dynamic_background}. In this section, we compare the performance of dynamic background subtraction techniques for the on-shore videos in Singapore Marine dataset. We have used the on-shore videos only to ensure that the dynamics correspond to the scene only and not to the sensor. The metrics and methodology of comparison are the same as discussed in section \ref{sec:quant_static_bkgnd}. Since the source codes or executables of the methods for maritime dynamic background subtraction are not available, with the exception of IMBS \cite{bloisi2014background}, we have used temporal mean (TempMean) background model implementation of \cite{bgslibrary} as an example of techniques that use the concept of relatively stationary pixels, adaptive median filtering (AdaMed) \cite{mcfarlane1995segmentation} as an example of spatio-temporal filtering approaches, Gaussian mixture model (GMM) of \cite{zivkovic2006efficient}, kernel density estimation (KDE) of \cite{elgammal2000non}, Lucas-Kanade approach \cite{lucas1981iterative} for optical flow (OptFlow) based background subtraction, and IMBS as an example of multistep approaches for background subtraction. We used the computer vision toolbox of Matlab for optical flow segmentation using Lucas-Kanade \cite{lucas1981iterative} approach. The video was scaled down to 0.5 times its actual size in pixels before computing the optical flow. Bounding boxes with dimensions less than 10 pixels in the scaled down frames were filtered away to suppress the motion due to water. The remaining bounding boxes were used after scaling up to the original dimensions. For IMBS \cite{bloisi2014background}, we have used the source code of the authors. For the rest, we have used the codes in the background subtraction library \cite{bgslibrary}.

The comparison results are shown in Table \ref{tab:quant_bck}. With the exception of temporal mean approach, the other dynamic background subtraction approaches invariably perform better than the static background subtraction approaches. Further, the noticeably better precision of optical flow based approach is attributed to the down-scaling of the frames which suppressed detection of extremely small spurious characteristic of motion of water and the filtering away of small foreground segmentations. In terms of recall, the adaptive median approach of \cite{mcfarlane1995segmentation} performs the best, despite being simple. Nevertheless, none of the methods provide practically useful precision and recall.

\vspace{-2mm}
\subsection{Tracking}

In a typical object tracking pipeline, objects are extracted by background subtraction and segmented before they are tracked. However, in some cases, tracking is done even without segmenting the background, as discussed in section \ref{subsec:seg_with_track}.

\subsubsection{Basic tracking techniques}\label{subsubsec:basictracking}
Hu et. al \cite{hu2011robust} formulated the problems of tracking as computation of an adaptive bounding box, where the bounding box in current frame is an adaptation of the bounding box in the previous frame within specified ranges of adaptivity to compensate for the background mismatch between the current and previous frames. Temporal high pass filter (analogous to fast moving objects) of segmented shapes was used in \cite{sumimoto1994machine}. Robert-Inacio et. al \cite{robert2007multispectral} tested the locations of the objects in consecutive frames for expected speed range. Objects were tracked as long as such speed of the object persists. Westall et. al \cite{westall2008evaluation} used dynamic programming for tracking of the objects in the videos.

\subsubsection{Feature based tracking}\label{subsubsec:featuretracking}
Methods in which prominent features of the objects are used for tracking are more suitable for dealing with occlusion. Bloisi et. al \cite{bloisi2011automatic} used Haar features for detecting and tracking objects. It is notable that \cite{bloisi2011automatic} used visible range color images as input and their features detection strategy for color images may not be directly useful for IR images. Other key point detectors, such as Harris corner detector, were found to be more effective \cite{van2012persistent}.

\begin{table*}
  \caption{Summary of tracking approaches used in maritime object tracking.}\label{tab:tracking}\vspace{-2mm}
  \centering
  \begin{tabular}{|p{1.85cm}||p{4.35cm}|p{3.1cm}|p{3.1cm}|p{3.5cm}|}
    \hline
    \multicolumn{1}{|c||}{Approach} & \multicolumn{1}{c|}{Model} & \multicolumn{1}{c|}{Learning} & \multicolumn{1}{c|}{Advantages} & \multicolumn{1}{c|}{Disadvantages} \\
    \hline
    \hline
    Basic tracking techniques \cite{hu2011robust,sumimoto1994machine,robert2007multispectral,westall2008evaluation} & Adaptive bounding box, Temporal high pass filter & Online using a small temporal window and memory of previous tracking & Simple, computation efficient, adaptive bounding box can deal with wakes & Naive, non-predictive\\
    \hline
    Feature based tracking \cite{bloisi2011automatic,van2012persistent} & Track features of segmented objects & Match features across frames & More robust than shape tracking, may allow some deformation (aspect change) & Features may not be consistently present, selection of appropriate features is important, computation intensive\\
    \hline
    Shape tracking with level-sets \cite{szpak2011maritime,frost2013detection} & Segmented shape contours are cast into level sets & Level sets are evolved through the frame & Apriori knowledge not required, but beneficial & Cannot deal with occlusion, sensitive to shape segmentation, computation intensive\\
    \hline
    Bayesian predictive network \cite{westall2008evaluation} & Features of objects case as state vectors of network representing motion model & Learning techniques such as expectation maximization are used to progressively update the network & Predictive nature, learns motion adaptively & Requires features as state variables, very computation intensive, cannot deal with complex motion patterns\\
    \hline
    Kalman filters \cite{bloisi2009argos,fefilatyev2010tracking,wei2009automated,fefilatyev2012detection,angelova2008extended} & Motion model is represented as state vector at a time t, current foreground segmentation's feature is case as input vector (to update the state vector) and Gaussian random motion perturbation is cast as noise (to be filtered) & State vector is updated for least square error between the actual measurement (input) and the predicted measurement (using previous state vector) & Almost real time, can filter away random perturbations, can adapt to multiple objects, can deal with complex motion variations occurring slowly & Does not work for complex random small motions, needs special framework for multiple object tracking \\
    \hline
    Optical flow \cite{bloisi2009argos,mittal2004motion} & Computes motion maps or flow vectors to determine consistent motion patterns & Cluster using modified k-means clustering on flow maps or compute flow equation at each pixel & Wakes and waves are automatically suppressed due to inconsistent motion patterns & Cannot deal with boat maneuvers and is computation intensive \\
    \hline
  \end{tabular}\vspace{-4mm}
\end{table*}

\subsubsection{Shape tracking with level-sets}
Casting segmented shapes (shape contours) as level-sets \cite{frost2013detection,szpak2011maritime} and then evolving the level-sets over frames has also been found useful for tracking. Notably, level-set techniques generally require the number of foreground objects and their initial contours to be specified \cite{frost2013detection,szpak2011maritime}. On one hand, knowing the number of objects require pre-segmentation of the objects, even if crude estimates are used. On the other hand, specifying initial contours imply that occluded objects may be difficult to deal with in such techniques. The level-set based approaches can benefit from some shape prior \cite{frost2013detection} which may be known through the general geometry knowledge of the expected objects (sea vessels of various kinds for maritime problem).

\subsubsection{Bayesian predictive network}\label{subsubsec:Bayesiantracking}

In the Bayesian approaches for tracking, the objects or features in a frame to be tracked are considered as a state vector of the Bayesian network and the training of a predictive model for transition between states is done to obtain a predictive model \cite{westall2008evaluation}. Often, once a Bayesian network is trained, it can be used to predict the state of the next frame given the state(s) in the previous frame(s). Thus Bayesian networks and hidden Markov models are suitable for tracking of objects that have a relatively smooth or predictable motion. They are suitable for dealing with multiple objects as well as occlusion as the objects' locations and features in a frame can be assigned a state variable each while the occlusion can be dealt with due to the predictive nature of the networks. However, it is difficult to make them adaptive and thus agile to learn complex motion characteristics such as some objects becoming stationary for some time.

\subsubsection{Kalman filters}

Tracking large objects such as ships in port environment was performed using a mixture Kalman filter approach in \cite{angelova2008extended}. Kalman filters were used in \cite{zhong2003segmenting,koller1994robust} for learning the motion of the foreground objects. In the original form, Kalman filters cannot deal with multiple hypotheses, or multiple object motion tracking simultaneously \cite{isard1996contour}. Thus, a multi-hypotheses Kalman filter for tracking was proposed in \cite{reid1979algorithm} and its optimal implementation was presented in \cite{cox1996efficient}. It was found useful in maritime problems \cite{bloisi2009argos,fefilatyev2010tracking,wei2009automated,fefilatyev2012detection}.

It was reported in \cite{bloisi2009argos} that the multi-hypotheses Kalman filter approach provides a good balance between computation load and tracking robustness. We note that Bloisi and Iocchi \cite{bloisi2009argos} tested this on high resolution video stream and its validity on lower quality videos is not assured. Wei et. al \cite{wei2009automated} used manually pre-assigned initial tracks and simple Kalman filters to track the objects instead of multi-hypotheses Kalman filters.

\subsubsection{Motion segmentation using optical flow approach}\label{subsec:seg_with_track}

Direct foreground tracking without pre-segmenting the foreground can be done by 'motion segmentation'. The spatial information is incorporated implicitly as the features or pixels with same motion characteristics are likely to belong to the same foreground object. Although several motion segmentation approaches are used in the computer vision, as discussed later in section \ref{subsec:track_CV}, methods for maritime EO problem have used optical flow based motion segmentation only \cite{bloisi2009argos,mittal2004motion}. The underlying assumptions in optical flow based techniques are that the objects are rigid and the motion is smooth.

Bloisi et. al \cite{bloisi2009argos} first computed connected segments in the foreground, which are not necessarily the foreground objects. Then, they computed sparse motion maps for each blob. The wakes and shadows do not have a consistent motion map and thus are suppressed in the optical flow approach. In order to deal with multiple foreground objects in one blob, a modified k-means clustering approach was applied on the optical flow map. First the motion map was over-clustered into many small clusters using k-means clustering. Then, the clusters were iteratively merged till further merging reduced the cluster separation instead of increasing the cluster separation. Notably, optical flow method fails in boat maneuvers because the optical flow may detect different directions for the different parts of the boat. Further, two boats having very similar motion characteristics and present in one blob cannot be separated by optical flow as well as k-means clustering.

Mittal et. al \cite{mittal2004motion} computed the optical flow velocity vector $f$ by solving the flow constraint equation $\nabla g \cdot f + g_t=0$, where $g$ is the measured value of a color channel, $\nabla g$ is the gradient of $g$, and $g_t$ is the temporal derivative of $g$. This equation was solved at each pixel such that the error in the estimated flow vector $f$ is minimised while satisfying the constraint that the velocity $f$ is locally constant.

\subsubsection{Comparison of techniques for tracking}

A qualitative comparison is given in Table \ref{tab:tracking}. Here, we present a performance comparison of techniques of tracking on on-shore videos of Singapore Marine dataset. We have used only on-shore videos so that the performance of tracking is not biased due to camera's motion. Performance metrics for tracking \cite{bernardin2008evaluating}, namely precision, recall, multiple object tracking accuracy (MOTA), multiple object tracking precision (MOTP), and false alarm rate (FAR) are used, which we describe below.

An $i$th ground truth (GT) track is represented by its GT bounding box $O^{\rm GT}_{i,t}$ in frame $t$. Analogously, $O^{\rm det}_{j,t}$ denotes the bounding box of the $j$th track at time $t$ detected by a tracking technique (simply referred to as a tracker). Then, for $i$th ground truth track and $j$th detected track, ${\rm IOU}(i,j,t)$ denotes the value of IOU computed using eq. (\ref{eq:IOU}) for the pair ($O^{\rm GT}_{i,t}$,$O^{\rm det}_{j,t}$) at frame $t$. Matched pairs of ground truth tracks and detected tracks are determined using Hungarian method \cite{kuhn1955hungarian} with $1-r(i,j,t)$ as input, where $r(i,j,t)$ is defined as
\begin{equation}\label{eq:r(i,j)}
  r(i,j,t) = \left\{ {\begin{array}{*{20}{c}}
{{\rm{IOU}}(i,j,t)}&{{\rm{if \,\, IOU}}(i,j,t) > 0.5}\\
0&{{\rm{otherwise}}}
\end{array}} \right.
\end{equation}
Hereafter, ($i,j$) denotes a matched pair of $i$th ground truth track and its corresponding $j$th detected track. In a frame $t$, all unmatched ground truth tracks contribute to one false negative (FN) each and all unmatched detected tracks contribute to one false positive (FP) each. For the matched tracks, if ${\rm IOU}(i,j,t)$ is more than 0.5 (based on \cite{bernardin2008evaluating}), the detection is said to be true positive (TP) for that frame. Otherwise it contributes to a mismatch (MM). Thus, $N_{{\rm TP},t}$, $N_{{\rm MM},t}$ , $N_{{\rm FP},t}$, and $N_{{\rm FN},t}$ are the numbers of TPs, mismatches, FPs, and FNs in a frame $t$. Further, total number of matched pairs of ground truth and detected tracks in a frame $t$ irrespective of the values of IOU is given as $N_{{\rm M},t} = N_{{\rm TP},t} + N_{{\rm MM},t}$ and the total number of frames in the video is denoted as $T$. Precision, recall, FAR, MOTA, and MOTP are then defined as
\begin{eqnarray}
  {\rm Precision} &=& \frac{\sum_t{N_{{\rm M},t}}}{\sum_t{\left({ N_{{\rm M},t} + N_{{\rm FP},t}}\right)}} \\
  {\rm Recall} &=& \frac{\sum_t{N_{{\rm M},t} }}{\sum_t{\left({ N_{{\rm M},t} + N_{{\rm FN},t}}\right)}} \\
  {\rm{FAR}} &=& {{\sum\limits_t {{N_{{\rm{FP}},t}}} }}/{T} \\
  {\rm MOTA} &=& 1-\frac{\sum_t{\left({ N_{{\rm FP},t} + N_{{\rm FN},t} + N_{{\rm MM},t} }\right)}}{\sum_t{\left({ N_{{\rm M},t} + N_{{\rm FN},t}}\right)}} \\
  {\rm{MOTP}} &=& \frac{{\sum\limits_{(i,j),t} {r\left( {(i,j),t} \right)} }}{{\sum\limits_t {{N_{{\rm{M}},t}}} }}
\end{eqnarray}
Precision and recall have their usual range of [0,1] with best values being 1. The unit of FAR is number of false positives per frame and its value may be any non-negative real number, with 0 being the best value. MOTA may take negative values but has a maximum and best value of 1. MOTP lies in the range [0,1], the best value being 1.

\begin{table}
  \centering
    \caption{Quantitative evaluation of different tracking techniques for on-shore videos of Singapore Marine dataset. The best values for each metric are highlighted using bold font.}\label{tab:quant_track}\vspace{-2mm}
  \begin{tabular}{|p{0.9cm}|p{0.5cm}||c|c|c|c|c|}
    \hline
    \multicolumn{7}{|c|}{Techniques related to maritime}\\
    \hline
    {Metric}&{}&MST&KLT&DAOT&MOT&LKDoG\\
    \hline
    \multirow{5}{*}{Precision} & {Mean} & 0.57 & \textbf{0.73} & 0.65 & 0.01 & 0.00\\
    & {Q25} & {0.38} & \textbf{0.60} & {0.53} & {0.00} & {0.00}\\
    & {Q50} & {0.52} & \textbf{0.70} & {0.65} & {0.01} & {0.00}\\
    & {Q75} & {0.83} & \textbf{0.86} & {0.77} & {0.01} & {0.00}\\
    \hline
    \multirow{5}{*}{Recall} & {Mean} & 0.53 & \textbf{0.77} & 0.68 & 0.04 & 0.02\\
    {} & {Q25} & {0.31} & \textbf{0.68} & {0.57} & {0.0.00} & {0.00}\\
    {} & {Q50} & {0.50} & \textbf{0.76} & {0.68} & {0.03} & {0.01}\\
    {} & {Q75} & {0.77} & \textbf{0.88} & {0.82} & {0.06} & {0.02}\\
    \hline
    \multirow{5}{*}{MOTA} & {Mean} & 0.15 & \textbf{0.47} & 0.29 & -6.45 & -11.79\\
    {} & {Q25} & {-0.20} & \textbf{0.20} & {0.07} & {-8.36} & {-10.02}\\
    {} & {Q50} & {0.05} & \textbf{0.45} & {0.32} & {-5.43} & {-8.00}\\
    {} & {Q75} & {0.59} & \textbf{0.72} & {0.59} & {-4.07} & {-5.67}\\
    \hline
    \multirow{5}{*}{MOTP} & {Mean} & 0.74 & \textbf{0.80} & 0.68 & 0.61 & 0.45\\
    {} & {Q25} & {0.71} & \textbf{0.76} & {0.66} & {0.59} & {0.51}\\
    {} & {Q50} & {0.74} & \textbf{0.81} & {0.68} & {0.60} & {0.55}\\
    {} & {Q75} & {0.77} & \textbf{0.82} & {0.69} & {0.62} & {0.59}\\
    \hline
    \multirow{5}{*}{FAR} & {Mean} & 3.46 & \textbf{2.70} & 3.56 & 46.85 & 81.10\\
    {} & {Q25} & {0.97} & \textbf{0.91} & {1.16} & {42.21} & {46.45}\\
    {} & {Q50} & {2.66} & \textbf{2.43} & {2.83} & {48.73} & {58.36}\\
    {} & {Q75} & {5.79} & \textbf{4.40} & {5.20} & {52.43} & {81.17}\\
    \hline
    \multicolumn{2}{|l||}{Time/frame (s)} & {1.66} & {0.51} & {0.73} &{\textbf{0.26}} & {1.14}\\
    \hline
    \hline
    \multicolumn{7}{|c|}{Other computer vision techniques}\\
    \hline
    {Metric}&{}& \scriptsize{AdaBoost} &MIL&TLD&MedFlow&KCF\\
    \hline
    \multirow{5}{*}{Precision} & {Mean} & 0.82 & 0.62 & 0.26 & 0.76 & \textbf{0.86}\\
    & {Q25} & 0.68 & 0.48 & 0.16 & 0.69 & \textbf{0.77}\\
    & {Q50} & 0.86 & 0.71 & 0.26 & 0.75 & \textbf{0.90}\\
    & {Q75} & 0.91 & 0.78 & 0.35 & 0.86 & \textbf{0.96}\\
    \hline
    \multirow{5}{*}{Recall} & {Mean} & 0.83 & 0.63 & 0.27 & 0.77 & \textbf{0.87}\\
    {} & {Q25} & 0.75 & 0.48 & 0.16 & 0.72 & \textbf{0.79}\\
    {} & {Q50} & 0.85 & 0.70 & 0.27 & 0.79 & \textbf{0.90}\\
    {} & {Q75} & 0.92 & 0.79 & 0.36 & 0.84 & \textbf{0.95}\\
    \hline
    \multirow{5}{*}{MOTA} & {Mean} & 0.64 & 0.23 & -0.50 & 0.52 & \textbf{0.72}\\
    {} & {Q25} & 0.46 & -0.03 & -0.71 & 0.39 & \textbf{0.55} \\
    {} & {Q50} & 0.66 & 0.42 & -0.49 & 0.52 & \textbf{0.80}\\
    {} & {Q75} & 0.83 & 0.54 & -0.30 & 0.68 & \textbf{0.89}\\
    \hline
    \multirow{5}{*}{MOTP} & {Mean} & 0.79 & 0.75 & 0.63 & 0.79 & \textbf{0.80}\\
    {} & {Q25} & 0.77 & 0.69 & 0.60 & 0.76 & \textbf{0.78}\\
    {} & {Q50} & \textbf{0.80} & 0.75 & 0.63 & 0.79 & \textbf{0.80} \\
    {} & {Q75} & 0.82 & 0.80 & 0.65 & \textbf{0.83} & \textbf{0.83}\\
    \hline
    \multirow{5}{*}{FAR} & {Mean} & 1.73 & 3.52 & 6.56 & 2.24 & \textbf{1.37}\\
    {} & {Q25} & 0.70 & 1.51 & 4.07 & 1.17 & \textbf{0.19}\\
    {} & {Q50} & 1.13 & 2.45 & 6.21 & 1.57 & \textbf{0.91}\\
    {} & {Q75} & 2.37 & 4.99 & 8.53 & 3.35 & \textbf{2.09}\\
    \hline
    \multicolumn{2}{|l||}{Time/frame (s)} & {12.03} & {3.31} & {42.73} & \textbf{0.42} & {0.47}\\
    \hline
  \end{tabular}\vspace{-4mm}
\end{table}

\begin{table}
  \caption{Summary of literature on object detection and tracking in maritime scenario.}\label{tab:summary_existing}\vspace{-2mm}
  \centering
  \scalebox{0.84}{
  \begin{tabular}{|l|c||c|c||c|c||c|c|c||c|c|c|c|}
    \hline
    \multirow{3}{*}{Articles} & \multirow{3}{*}{Year} & \multicolumn{2}{c||}{{EO}} & \multicolumn{2}{c||}{{Scene}}  & \multicolumn{3}{|c||}{{Object}} & \multicolumn{4}{c|}{{Object}}\\
    {} & {} & \multicolumn{2}{c||}{{sensor}} & \multicolumn{2}{c||}{{}}  & \multicolumn{3}{|c||}{detection} & \multicolumn{4}{c|}{{tracking}}\\
    \cline{3-13}
     & &  \rotatebox{90}{Infrared}  & \rotatebox{90}{Visible} & \rotatebox{90}{On shore} & \hspace{-0.25cm} \rotatebox{90}{Open sea} & \rotatebox{90}{Horizon detection} &  \rotatebox{90}{Static background} & \rotatebox{90}{Foreground segmentation} & \rotatebox{90}{Horizon detection} &  \rotatebox{90}{Registration}   & \rotatebox{90}{Dynamic background} &   \rotatebox{90}{Foreground tracking} \\
    \hline
    \hline
    Bhanu \cite{bhanu1990model} & 1990
    & \textbullet & \textbullet
    & &
    & & \textbullet &
    & & & &
    \\
    \hline
    Sumimoto \cite{sumimoto1994machine} & 1994
    & & \textbullet
    & &
    & & & \textbullet
    & & & &
    \\
    \hline
    Strickland \cite{strickland1997wavelet} & 1997
    & & \textbullet
    & &
    & & &
    & & & \textbullet &
    \\
    \hline
    Smith \cite{smith1999identification} & 1999
    & \textbullet &
    & &
    & & \textbullet &
    & & & &
    \\
    \hline
    Broek \cite{van2000detection} & 2000
    & \textbullet &
    & &
    & \textbullet &\textbullet & \textbullet
    & & & &
    \\
    \hline
    Voles \cite{voles2000nautical} & 2000
    & &
    & &
    &  & {} & {}
    & {\textbullet} & {} & {\textbullet} & {\textbullet}
    \\
    \specialrule{.15em}{.075em}{0.075em}
    Caspi \cite{caspi2002spatio} & 2002
    &\textbullet &\textbullet
    &\textbullet&
    & & &
    & & \textbullet & & \textbullet
    \\
    \hline
    Ablavsky \cite{ablavsky2003background} & 2003
    & &
    & &
    & & &
    & & & \textbullet & \textbullet
    \\
    \hline
    Mittal \cite{mittal2004motion} & 2004
    & & \textbullet
    & &
    & & &
    & & & \textbullet & \textbullet
    \\
    \hline
    Socek \cite{socek2005hybrid} & 2005
    & & \textbullet
    & &
    & {} & {\textbullet} & {\textbullet}
    & & & {\textbullet} & {}
    \\
    \specialrule{.15em}{.075em}{0.075em}
    Fefilatyev \cite{fefilatyev2006horizon} & 2006
    & & \textbullet
    & \textbullet & \textbullet
    & \textbullet & &
    & & & &
    \\
    \hline
    Wang \cite{Wang2006SVMWavelets} & 2006
    &\textbullet &
    & &
    & & & \textbullet
    & & & \textbullet &
    \\
    \specialrule{.15em}{.075em}{0.075em}
    Robert- & 2007
    &\textbullet &\textbullet
    &\textbullet &
    &&\textbullet&
    &&&&\textbullet
    \\
    Inacio \cite{robert2007multispectral} & {}
    & &
    & &
    & & &
    & & & &\\
    \hline
    Schwering \cite{schwering2007eo} & 2007
    & \textbullet  & \textbullet
    & \textbullet &
    & & &
    & & & &
    \\
    \specialrule{.15em}{.075em}{0.075em}
    Bouma \cite{bouma2008automatic} & 2008
    &\textbullet &
    & \textbullet & \textbullet
    & \textbullet     & \textbullet & \textbullet
    & & & &
    \\
    \hline
    Broek \cite{van2008discriminating} & 2008
    & \textbullet & \textbullet
    & &
    & {} & {\textbullet} & {}
    & {} & {} & {} &
    \\
    \hline
    Zheng \cite{Zheng2008Mutlisensory} & 2008
    & \textbullet & \textbullet
    &&
    &&&
    &&&&
    \\
    \specialrule{.15em}{.075em}{0.075em}
    Bloisi \cite{bloisi2009argos} & 2009
    & & \textbullet
    & &
    & & &
    & & & \textbullet & \textbullet
    \\
    \hline
    Gupta \cite{gupta2009adaptive} & 2009
    & &
    & &
    & & &
    & \textbullet & & \textbullet & \textbullet
    \\
    \hline
    Haarst \cite{van2009polynomial} & 2009
    & & \textbullet
    & &
    & & \textbullet &
    & & & &
    \\
    \hline
    Wei \cite{wei2009automated} & 2009
    & & \textbullet
    & &
    & \textbullet & \textbullet & \textbullet
    & & & &
    \\
    \specialrule{.15em}{.075em}{0.075em}
    Fefilatyev \cite{fefilatyev2010tracking} & 2010
    & & \textbullet
    & \textbullet & \textbullet
    & & \textbullet &
    & \textbullet & \textbullet & & \textbullet
    \\
    \hline
    Zhu \cite{zhu2010novel} & 2010
    & &
    & &
    & & \textbullet &
    & & & \textbullet&
    \\
    \specialrule{.15em}{.075em}{0.075em}
    Bloisi \cite{bloisi2011automatic} & 2011
    & &\textbullet
    & \textbullet &
    & {\textbullet} & &
    & & & &
    \\
    \hline
    Hu \cite{hu2011robust} & 2011
    & & \textbullet
    & \textbullet &
    & & &
    & & & \textbullet & \textbullet
    \\
    \hline
    Szpak \cite{szpak2011maritime} & 2011
    & & \textbullet
    & &
    & {} & {\textbullet} & {\textbullet}
    & {} & {} & {} &
    \\
    \hline
    Wang \cite{Wang2011FuzzyClutter} & 2011
    &\textbullet&
    &&
    &\textbullet &\textbullet&
    &&& &
    \\
    \specialrule{.15em}{.075em}{0.075em}
    Broek \cite{van2012persistent} & 2012
    & \textbullet & \textbullet
    & \textbullet &
    &&& \textbullet
    &&&&
    \\
    \hline
    Ren \cite{Ren2012} & 2012
    & &\textbullet
    & &
    & & &
    & & & \textbullet &
    \\
    \hline
    Zhang \cite{zhang2012visual} & 2012
    &&\textbullet
    &\textbullet&
    & &\textbullet &
    &&&&
    \\
    \specialrule{.15em}{.075em}{0.075em}
    Frost \cite{frost2013detection} & 2013
    & &
    & &
    & & \textbullet &
    & & & & \textbullet
    \\
    \hline
    Gershikov \cite{gershikov2013horizon} & 2013
    & \textbullet & \textbullet
    & &
    & \textbullet & &
    & & & &
    \\
    \hline
    Tang \cite{tang2013research} & 2013
    &\textbullet &
    & &
    & \textbullet & \textbullet &
    & & & &
    \\
    \specialrule{.15em}{.075em}{0.075em}
    Bloisi \cite{bloisi2014background} & 2014
    & & \textbullet
    & &
    & & &
    & & & \textbullet &
    \\
    \hline
    Broek \cite{van2014ship} & 2014
    & \textbullet &
    & &
    & \textbullet &\textbullet & \textbullet
    & & & &
    \\
    \hline
    Broek \cite{van2014recognition} & 2014
    & \textbullet &
    & &
    & \textbullet &\textbullet & \textbullet
    & & & &
    \\
    \hline
    Chen \cite{Chen2014} & 2014
    & \textbullet &
    & &
    & & \textbullet & \textbullet
    & & & &
    \\
    \hline
    Tu \cite{tu2014infrared} & 2014
    & \textbullet &
    & &
    & & &
    & & & \textbullet &
    \\
    \hline
    Wang \cite{wang2014aquatic} & 2014
    & & \textbullet
    & &
    & \textbullet & \textbullet &
    & & \textbullet & &
    \\
    \hline
    Zhou \cite{Zhou2014} & 2014
    & & \textbullet
    & &
    & & \textbullet &
    & & & &
    \\
    \specialrule{.15em}{.075em}{0.075em}
    Babaee \cite{babaee20153} & 2015
    & & \textbullet
    & &
    & & &
    & & \textbullet & &
    \\
    \hline
    Wang \cite{wang2015aquatic} & 2015
    & & \textbullet
    & &
    & & &
    & & & & \textbullet
    \\
    \hline
  \end{tabular}}
  \vspace{-4mm}
\end{table}

We consider one technique per row of Table \ref{tab:tracking}, with the exception of level-set based tracking, for which we could not find an implementation. We use mean shift tracking\footnote{{\url{https://www.mathworks.com/matlabcentral/fileexchange/35520-mean-shift-video-tracking}}} (MST) \cite{bradski1998computer} as an example of basic tracking techniques, Kanade-Lucas-Tomasi (KLT) feature tracker\footnote{\label{note:matlab}Matlab provided function} \cite{lucas1981iterative,tomasi1991detection} for feature based tracking, distractor-aware online tracking (DAOT\footnote{Matlab code provided by the authors of \cite{possegger2015defense})}, \cite{possegger2015defense}) for tracking using Bayesian predictive network, motion-based multiple object tracking\textsuperscript{\ref{note:matlab}} (MOT) using Kalman filter \cite{li2010multiple} for Kalman filter based tracking, and optical flow based on Lukas-Kanade difference of Gaussian method\textsuperscript{\ref{note:matlab}} (LKDoG, \cite{barron1992fleet}). No background subtraction technique has been applied in order to compare the performance of tracking only. MST, KLT, and DAOT are single object trackers and require initial guess (we used the first bounding box of each GT track). MOT and LKDoG do not require initial guess and track multiple object simultaneously. The results are presented in Table \ref{tab:quant_track}. MST, KLT, and DAOT clearly benefit from the initial guess because the number of tracks remains close to the actual number of ground truth tracks. On the other hand, MOT and LKDoG suffer due to large number of false positives as a consequence of water dynamics. Among MST, KLT, and DAOT, MST performed the best in terms of all the metrics.

\vspace{-2mm}
\section{Computer vision approaches beyond maritime}\label{sec:computer_vision_only}

A literature summary of \textbf{maritime} EO data processing for object detection and tracking is provided in Table \ref{tab:summary_existing}. Object detection and tracking has been studied for several decades in computer vision as well. However, due to the specific set of challenges presented by the maritime environment, not much attention has been paid in developing algorithms specific to this domain. Nevertheless, given the large number of algorithms already developed for object detection and tracking over the past years, it is only natural to seek out the algorithms that may be suitable in the maritime scenario. In this section, we identify some such algorithms which have reported at least one example with dynamic water background.

\vspace{-2mm}
\subsection{Object detection}

In the context of current maritime EO data processing, the foreground obtained after background subtraction is segmented and the segmented regions are identified as the objects of interest. Thus, object detection is mainly performed through background subtraction. Different approaches have been taken for modelling and segmentation of background. There are several useful surveys on the topic of background suppression in video sequences \cite{elhabian2008moving}. Below, we discuss the techniques that are potentially effective in maritime videos.

\subsubsection{Relatively stationary pixels}

In the context of the section \ref{subsubsec:relative}, we discuss other relevant works from non-maritime applications. Weighted average of intensities at a pixel across time was considered in \cite{cavallaro2000video,koller1994towards}. Median filter was employed for background suppression in \cite{Sang1998,Barnett1989,cucchiara2003detecting,el2007outdoor,shoushtarian2005practical}.
First order low pass filtering was used in \cite{el2007outdoor}. Toyama et. al \cite{toyama1999wallflower} used pixel-wise temporal filter (Wiener filter). All the temporal filters essentially use temporal variation at pixels as indicator of foreground and background.

\subsubsection{Spatio-temporal filtering}
Ridder et. al \cite{ridder1995adaptive} proposed to use Kalman filter for background estimation. This approach was found to be robust to illumination changes and incorporated pixel-wise automatic threshold (thus was less sensitive to control parameters). Zhong and Sclaroff \cite{zhong2003segmenting} also used Kalman filter for representing dynamic textures.

\subsubsection{Gaussian Mixture models}
While initial forms of Gaussian mixture models have already found use in maritime background subtraction \cite{bloisi2009argos,wang2014aquatic,fefilatyev2012detection,frost2013detection,gupta2009adaptive}, GMM has also been increasingly combined with other techniques in the computer vision community to improve the performance of object detection specifically in challenging dynamic environments. For example, the local variation persistence method \cite{Pham2015332} uses GMM for separating static background as the Gaussian component with large standard deviation and removing it, followed by numerical computation of negative differential entropy of the remaining Gaussian components which allowed for separating locally persistent variations as dynamic background. Varadarajan et. al \cite{Varadarajan20153488} propose to use a square region based GMM, which inherently considers local spatial variations in addition to temporal variations in order to obtain a better background model for challenging dynamic backgrounds including water bodies.

\subsubsection{Kernel density estimation}

As mentioned in section \ref{subsubsec:KDE}, suitable kernels can be chosen for the KDE model of background. Chen and Meer \cite{chen2002robust} proposed to use Epanechnikov kernels \cite{epanechnikov1969non}, which is optimal in the least square error sense. Kato et. al \cite{kato2002hmm} used a Gaussian distribution for intensity variation at a pixel, a Gaussian distribution for wavelet coefficient variation at a pixel, and their combination as a single 2-dimensional Gaussian kernel. An adaptive scheme for KDE model update was proposed in \cite{zivkovic2006efficient}, where the volumes (spreads) of the kernels were made adaptive by changing the number of frames considered for the dynamic update.

\subsubsection{Optical flow}
Ross \cite{ross2000exploiting} presented an interesting concept of texture-and-motion duality in optical flow in order to extract background. It used the single image segmentation approach of \cite{felzenszwalb2004efficient} to get an initial estimate of the background. Optical flow of the segmented regions was computed using an energy minimization approach \cite{horn1981determining}. Li and Xu \cite{Li201684} perform optical flow computation directions at the edges of super-pixelated regions to enhance the computation speed while allowing the identification of super-pixelated regions belonging to the dynamic background owing to the non-uniform flow vectors at their edges.

\begin{table*}
  \caption{Background subtraction approaches in computer vision having potential in maritime scenario (not covered in Tables \ref{tab:static_background} and \ref{tab:dynamic_background}).} \label{tab:comp_vis_background}\vspace{-2mm}
  \centering
  \begin{tabular}{|p{3cm}||p{4.5cm}|p{5cm}|p{4cm}|}
    \hline
    \multicolumn{1}{|c||}{Approach} & \multicolumn{1}{c|}{Description} & \multicolumn{1}{c|}{Advantages} & \multicolumn{1}{c|}{Disadvantages} \\
    \hline
    \hline
    {Range model \cite{haritaoglu2000w,kim2004background}} & {Multiple class model, each with a range of intensities} & {Simple, pre-learnt ranges, may be made adaptive} & {Not discriminative} \\ \hline
    {Local binary and ternary patterns \cite{ojala2002multiresolution}$-$\nocite{heikkila2006texture,shenbiological,zhong2013background,lin2014complex,tan2010enhanced,liao2010modeling,chan2009layered}\cite{chan2008modeling}} & {Background model representing dynamic textures} & {Multiple patterns may be learnt for different dynamic textures such as water, wake, and waves, quite robust, can be learnt and adapted online} & {Computation intensive} \\ \hline
    {Hidden Markov model \cite{socek2005hybrid},\cite{kato2002hmm}$-$\nocite{elhabian2008moving,ross2000exploiting,friedman1997image,rittscher2000probabilistic,cevher2009sparse,stenger2001topology,mumtaz2014joint}\cite{sheikh2005bayesian}} & {Local intensities (or other features) as state vectors in HMM} & {Uses temporal continuity of classification, does not require any pre-learning} & {Complex, computation intensive} \\ \hline
    {Saliency based approaches \cite{wixson2000detecting}$-$\nocite{itti1998model,gao2008plausibility}\cite{mahadevan2010spatiotemporal}}  & {Approaches for classifying pixels as background based on the used background model} & {More sophisticated than a simple threshold or range; can incorporate certain properties for classification, such as discriminative property and surprise} & {Complex, may be computation intensive} \\ \hline
    {Fuzzy classifiers \cite{xia2012segmentation}$-$\nocite{Xia2008BackgroundSuppression,Xia2011_BuildingDetection}\cite{chang2000applying}} & {Fuzzy techniques for classification of pixel as background} & {Needs appropriate fuzzy classifier functions to be pre-learnt} & {Complex, needs supervised learning of classifier functions} \\ \hline
    {Subspace based approaches \cite{wang2015aquatic,zhang2012visual,mason2001using,zhong2003segmenting,lin2014complex,cevher2009sparse,gao2008plausibility}, \cite{seki2003background}$-$\nocite{de2001robust,tsai2009independent,kim2011spatiotemporal,power2002understanding,oliver2000bayesian,toyama1999wallflower,monnet2003background,matsuyama2000background,guyon2013foreground,cevher2008compressive,dikmen2008robust,huang2009learning,mairal2010network}\cite{shen2012efficient}} & {Learning the background model, compactly representing and fast updating of the model, finding the overlap of pixel features with the model} & {Fast, compact, amenable to fast linear programming} & {Assume linear separability of data, degrade with large dynamics in background} \\
    \hline
  \end{tabular}\vspace{-4mm}
\end{table*}

\subsubsection{Range model}
A simple and popular approach for dynamic background extraction was considered in \cite{haritaoglu2000w} which used a range of intensity values for a given pixel, quantified by minimum and maximum intensity values at a background pixel and maximum intensity difference between two consecutive frames, denoted as $m(x),n(x),d(x)$, respectively. For finding the parameters of this model, the pixel's intensity values in a reasonably long time sequence $I(x,t)$ were used. First, the instances $t'$ at which pixel can be considered as stationary were found as
\begin{equation}\label{eq:stationarypixels}
|I(x,t')-\lambda(x)| < 2\sigma(x)
\end{equation}
\noindent where $\lambda(x)$ and $\sigma(x)$ are the mean and standard deviation of $I(x,t), \forall t$. Then, $m(x) = \min(I(x,t'), n(x) = \max(I(x,t'), d(x) = \max (|I(x,t') - I(x,t'-1)|)$ were computed. The model parameters may be updated as often as needed. Haritaoglu \cite{haritaoglu2000w} also suggested a technique for identifying that a moving object in earlier frames has become a stationary background in later frames.

Kim et. al \cite{kim2004background} used a codebook of possible range values for addressing multi-class background. The code of a class was given by the range parameters discussed above. This was further augmented by average color data, frequency of occurrence of the code, and last access of the code.

\subsubsection{Dynamic textures}\label{subsubsec:LBP}

Local binary pattern \cite{ojala2002multiresolution} (LBP), either at a single pixel, or a small region around the given pixel \cite{heikkila2006texture}, finds a binary number representing the boolean intensity changes in the neighborhood of the chosen pixel. It may be made shift and rotation invariant, as discussed in \cite{ojala2002multiresolution}. The LBP feature vector of a block of pixels in a frame is the histogram of the binary numbers obtained at all the pixels in the block \cite{heikkila2004texture}. Over time, one LBP feature vector is obtained for each frame and the net background feature vector is a weighted combination of feature vectors of the last $K$ (often heuristically chosen) number of frames. A distance measure for decision making and a model update scheme is discussed in \cite{heikkila2006texture}. This approach was found useful in applications involving underwater videos \cite{shenbiological,zhong2013background}. Local binary similarity patterns (LBSP) \cite{st2014improving} are a variation of LBP and include spatio-temporal binary similarity metric. A modification of LBP to deal with flat regions in an image is the local ternary patterns (LTP), presented and discussed in \cite{heikkila2006texture,lin2014complex,tan2010enhanced,liao2010modeling}. Furthermore, \cite{chan2009layered,chan2008modeling} proposed a mixture of dynamic textures, analogous to GMM, in order to allow for modeling of multiple dynamic textures. Mixture of dynamic textures showed good ability to deal with ocean's dynamic texture with synthetic translucent objects and flames.

\subsubsection{Hidden Markov model of dynamic background}

Hidden Markov models (HMM) have two specific advantages as compared to other modelling approaches \cite{kato2002hmm}. The first is its ability to incorporate temporal continuity. A pixel may be classified as belonging to background, foreground, or shadow in a particular frame. Nevertheless, it is likely that the pixel will have the same classification for at least a few continuous frames. This is so either because the object at the pixel is stationary or because the moving object occupies the pixels for some number of frames till the object crosses the pixel completely. HMM is inherently able to cover both these possibilities. Second, HMM does not require a specifically chosen training data. A scenario specific ordinary image sequence is sufficient for it to learn the hidden states that allow demarkation between background, foreground, and shadows. Further, it was noted in \cite{elhabian2008moving} that HMM approach is very effective in dealing with sudden illumination changes and providing a corrective temporary estimate of the background in such scenario.

Thus, despite being computationally expensive and difficult for dynamic modification of topology \cite{elhabian2008moving,ross2000exploiting}, HMM has attracted a lot of attention for background suppression \cite{socek2005hybrid,friedman1997image,rittscher2000probabilistic,cevher2009sparse,stenger2001topology,oliver2000bayesian}. One of the most recent works in this context is \cite{mumtaz2014joint}, which showed some examples of boats in sea as well. It has many interesting and useful features, which include using dynamic textures \cite{doretto2003dynamic} for simultaneous foreground-background modelling, augmenting the dynamic textures by introducing spatially smooth segmentation through HMM \cite{chan2009layered} and a specially designed expectation maximization approach with variational constraint.

We briefly discuss the update methods used for learning and updating the HMMs.
Sheikh and Shah \cite{sheikh2005bayesian} used Markov random field with maximum-a-posteriori estimation to obtain spatial context in a simultaneous foreground-background modeling approach. Expectation maximization approaches for training HMM have been discussed in \cite{friedman1997image,neal1998view,nowlan1991soft}. Stenger et. al \cite{stenger2001topology} designed a dynamic update scheme for HMM which allows for adaptive topology modification of the HMM. Ostendorf and Singer \cite{ostendorf1997hmm} suggested that dynamic adaptation of  HMM can be made fast by a state splitting approach. Brand and Kettnaker \cite{brand2000discovery} suggested that an arbitrarily large number of states may be initially chosen and then entropy based training of HMM may be used to identify the less probable states and iteratively remove them. Wang et. al \cite{wang2006dynamic} used an offline Baum Welch algorithm \cite{rabiner1989tutorial} to learn HMM but employed an online algorithm for background detection and updating the HMM. Rittscher at. al \cite{rittscher2000probabilistic} proposed a scheme for making HMM computationally less expensive and almost real-time. Brand and Kettnaker \cite{brand2000discovery} and Ostendorf and Singer \cite{ostendorf1997hmm} discussed the optimal choice of number of states of HMM. Further, some amount of speed up of the HMM update may be achieved using subspace based approaches \cite{zhong2003segmenting,cevher2009sparse}.

\subsubsection{Saliency based approaches for segmenting the background}

Wixson \cite{wixson2000detecting} used a saliency measure defined on the cumulative optical flow directions of the moving objects (foreground). It incorporates net flow directions by computing maximum flow directions and finding observations consistent with the maximum flow direction. Such saliency measure based on maximum flow direction may be suitable for single object tracking but may need significant modification for incorporating multiple object tracking. On the other hand, Itti et. al \cite{itti1998model} used a surprise based saliency map to segregate the non-surprising elements as the background (low saliency). It determined a surprising element as an element which has a large contrast compared to the surrounding pixels. The contrast should be consistently present at various length scales. This contrast is referred to as the center-surround difference. Although it can deal with the wavy nature of water to some extent, it is not effective in suppressing the wakes since they introduce a high contrast with respect to their surroundings.

Gao et. al \cite{gao2008plausibility} modified the saliency approach of \cite{itti1998model} by retaining the center-surround and multi-scaling. It used discrimination (referred to as discriminant in \cite{gao2008plausibility}) between intermediate features in the center-surround instead of using the direct contrast feature \cite{itti1998model} directly. It was tested on videos of floating bottle and surfer and showed better background identification than \cite{itti1998model}. We note that \cite{gao2008plausibility} used local ternary patterns as the features of the background model.

Mahadevan and Vasconcelos \cite{mahadevan2010spatiotemporal} combined the discriminant saliency approach of \cite{gao2008plausibility} and mixture of dynamic textures \cite{chan2008modeling} for determining spatially normal (high probability distributions) and temporally normal (high probability events) features of the videos with the moving crowd in urban scenarios. Furthering this concept, Wang et. al \cite{wang2006dynamic} proposed a saliency metric called spatiotemporal condition information (SCI). This metric computes the conditional information value (logarithm of conditional probability) of a pixel given the background and the spatiotemporal neighborhood of the pixel. Larger value of the conditional information indicates higher likelihood of the pixel being foreground.

Fang et. al \cite{fang2014video} computed two saliency maps, one characterizing spatial saliency through proximity and continuity of a visually salient object region and the other characterizing temporal saliency which accounts for dynamic background variation and persistence of local contrast. These maps are merged by using an adaptive entropy-based uncertainty weighing approach to form the final spatiotemporal saliency map.

Recently, Liu et. al \cite{liu2015background} used motion saliency map of \cite{xue2012motion} to determine the control parameter of the robust principle component analysis \cite{de2001robust}, which was then used for background subtraction and foreground extraction. In this definition of motion saliency, the sum of the background motion map $M$ such as due to water dynamics in maritime scenario and the stable background map $B$ is said to be a low rank representation of the video $V$. The background motion map and the stable background are solved by minimizing the sum of nuclear norm of $B$ and $L1$ norm of $M$.

\subsubsection{Fuzzy classification of background pixels}

Fuzzy logic was used to compute an adaptive threshold for classifying the background pixels in complex background in IR images \cite{xia2012segmentation,Xia2008BackgroundSuppression,Xia2011_BuildingDetection}. Although most of the experiments are in urban and semi-urban land scenarios, the techniques may inspire further interesting work in maritime IR images. A combination of fuzzy neural network and self organizing map \cite{maddalena2008self} was used in \cite{maddalena2010fuzzy}. It is shown to be robust to illumination changes and shadows, a property beneficial for maritime videos. In \cite{el2009fuzzy}, the usual GMM background model was modified to be fuzzy mixture of Gaussian functions.

\subsubsection{Subspace based approaches in background modelling and subtraction}

Spatio-temporal block of images, the collection of all the features of background model and the feature attributes of all the pixels may be represented as matrices. Then matrix decompositions can be used for manipulating the data, learning the model, compactly representing and fast updating the model, as well as finding the overlap of pixel features with the model in a powerful manner.

Thus, subspaces based approaches have been found useful \cite{vidal2010tutorial} for compact representation of features at pixels before learning is performed. These include eigenbackground approach \cite{zhang2012visual,sajid2014background,Sajid20154530,oliver2000bayesian}, principal component analysis (PCA) \cite{seki2003background,power2002understanding,oliver2000bayesian}, robust PCA \cite{de2001robust,Cao20161014}, independent component analysis \cite{tsai2009independent,Tu2015539}, and discriminant center surround \cite{gao2008plausibility,kim2011spatiotemporal}. Subspace based learning was used for night-time videos as well in \cite{zhao2008spatio}. It is anticipated that these approaches may be improved by suitable combination of eigenvectors of the data and designing a robust update scheme for the background.

Autoregressive updates of background models is done by identifying the subspace of consistently recurring backgrounds \cite{zhong2003segmenting,lin2014complex,toyama1999wallflower,monnet2003background}. Methods in \cite{mason2001using,matsuyama2000background,guyon2013foreground} that use correlation or covariance approaches may also be considered as subset of approaches that employ background subspace analyses. Some methods also use sparsity priors and implement background detection problem as sparse image reconstruction problem \cite{cevher2009sparse,cevher2008compressive,dikmen2008robust,huang2009learning,mairal2010network}. Alternatively, compressive sampling based approaches can be used to reduce dimensionality of the data before using other background detection techniques for reducing the computational cost \cite{wang2015aquatic,shen2012efficient}.

\begin{table*}
  \caption{Object tracking approaches in computer vision having potential in maritime scenario (not covered in Table \ref{tab:tracking}).}\label{tab:comp_vis_foreground}\vspace{-2mm}
  \centering
  \begin{tabular}{|p{2.75cm}||p{6.95cm}|p{3.85cm}|p{2.8cm}|}
    \hline
    \multicolumn{1}{|c||}{Approach} & \multicolumn{1}{c|}{Description} & \multicolumn{1}{c|}{Advantages} & \multicolumn{1}{c|}{Disadvantages} \\
    \hline
    \hline
    {Temporal persistence \cite{sheikh2005bayesian,grimson1998using}} & {Dynamic programming, progressively update motion by finding features/objects from previous segmentation} & {Fast update} & {Simple, non-predictive, not robust to occlusion} \\ \hline
    {Machine learning of motion \cite{nair2004unsupervised}$-$\nocite{grabner2006real,grabner2008semi,adam2006robust,babenko2011robust,okuma2004boosted,ross2008incremental}\cite{black1998eigentracking}} & {Learning techniques for learning motion patterns from segmented features/objects; depending upon the technique, may need pre-learning of patterns and matching them in frames, or off-line processing of a small subset or complete set of image frames for determining the motion characteristics } & {Robust to occlusion, provide complete motion characteristics} & {Non-predictive, complex, not real-time} \\ \hline
    {Optical flow \cite{mittal2004motion,brox2006variational,ablavsky2003background,wixson2000detecting,vidal2004unified}$-$\nocite{cremers2005motion,amiaz2006piecewise}\cite{black1996robust}} & {Motion maps learnt for segmented objects or just features in image frames} & {Can deal with occlusion, can use multiple features simultaneously} & {Computation intensive, many dense motion layers for complex motion} \\ \hline
    {Feature tracking and clustering \cite{vidal2004unified,tron2007benchmark}$-$\nocite{sheikh2009background,brox2010object}\cite{ochs2011object}} & {Features with same motion patterns are expected to belong to the same object} & {No explicit segmentation of object needed, very robust to occlusion, may be very discriminative} & {Not real-time, computation intensive} \\ \hline
    {Markov random field \cite{zhou2013moving}} & {Connected components network, where the features represent the state variables and all state variables may influence each other} & {Can deal with complex inter-dependent motion of multiple objects, robust to occlusions} & {Computation intensive, slow update} \\
    \hline
  \end{tabular}\vspace{-4mm}
\end{table*}

\subsubsection{Comparison of CV based background substraction techniques} A comparative summary of the background subtraction methods used in computer vision problems, but not covered in Tables \ref{tab:static_background} and \ref{tab:dynamic_background}, is given in Table \ref{tab:comp_vis_background}. Performance of five CV based techniques is compared in Table \ref{tab:quant_bck}. These techniques are LBP \cite{heikkila2006texture}, LBSP \cite{st2014improving}, FuzzGMM \cite{el2009fuzzy}, FAdaSOM \cite{maddalena2008self}, and Eigen \cite{oliver2000bayesian}. LBP and LBSP are dynamic texture approaches, FuzzyGGM is a fuzzy approach, AdaSOM is a neurofuzzy approach, and EigHMM uses a combination of HMM and eigenbackground (subspace based approach). The openCV implementations at the background subtraction library \cite{bgslibrary} are used for these methods. EigHMM performs the best in terms of recall and compares well with the recall values of AdaMed, showing good potential for maritime images. All methods perform poor in terms of precision, indicating large false positives. However, optimal choice of control parameters may render LBSP useful. Thus, we think that a combination of LBSP and eigen background may be helpful for maritime images.

\vspace{-2mm}
\subsection{Object tracking}\label{subsec:track_CV}

Here, we discuss the object tracking techniques developed for non-maritime situations, but hold promise for maritime scenarios. Sections \ref{subsubsec:foreground_CV} to \ref{subsubsec:machine} discuss tracking of segmented objects while the methods in sections \ref{subsubsec:optical_Flow} to \ref{subsubsec:MRF} do not need prior segmentation of objects.

\subsubsection{Foreground models}\label{subsubsec:foreground_CV}

Many methods represent the foreground objects by mixture models such as GMM, local ternary patterns (LTP), and KDE, similar to the dynamic background models. The components of the mixture components are used as the features \cite{chan2009layered,chan2008modeling,mumtaz2014joint}. This permits cushion for deformability, swivel, and small randomness in motion, which are useful for modelling moving sea vessels. Alternatively, the silhouette of the vessel may be tracked \cite{cremers2005motion,bertalmio2000morphing}. Greenberg et. al \cite{Greenberg2000} used morphological region growing and segmentation pruning after binarization for detecting objects with small false alarm rate.

A reverse approach was adopted in \cite{zhou2013moving,liu2015background}, inspired by \cite{candes2011robust}. Zhou et. al \cite{zhou2013moving} approximated the video as a low rank matrix and all the moving objects as systematic outliers to the low rank matrix belonging to an outlier support. Similar idea was used by Zhong and Sclaroff \cite{zhong2003segmenting} and Liu et. al \cite{liu2015background}, where foreground objects were considered as outliers to the background model corrupting the estimate of the background.

\subsubsection{Temporal persistence and dynamic programming for tracking}

A sophisticated version of dynamic programming approach is used in \cite{grimson1998using}, where dynamic programming updates the model parameters of the GMM representing the pre-segmented foreground object. A similar approach called temporal persistence was proposed in \cite{sheikh2005bayesian}, which assumed that a mobile foreground object would remain in spatial vicinity in consecutive frames and maintain similar color or intensity values. We note that the approach of \cite{sheikh2005bayesian} falls in motion segmentation category where pre-segmentation of the foreground is not assumed and the persistently mobile Gaussian mixtures over a few frames are concluded as foreground.

\subsubsection{Machine learning for tracking foreground objects}\label{subsubsec:machine}
Machine learning techniques such as boosting based unsupervised or semi-supervised boosting techniques are often used for tracking or motion learning \cite{nair2004unsupervised,grabner2006real,grabner2008semi,adam2006robust,babenko2011robust}.
Often, initialization through manual segmentation (such as in \cite{nair2004unsupervised,grabner2008semi}) is needed. Boosting approaches are quite useful since they can often deal with occlusion intrinsically by considering each object's motion independently \cite{okuma2004boosted} and learning them over a subset or all of the frames. When the subset of frames is used at a time, often online learning can be used \cite{grabner2006real,grabner2008semi,babenko2011robust}. Another method is the use of principal component analysis, where a low-dimensional subset of principal components is updated as new frames arrive \cite{ross2008incremental,black1998eigentracking}. Such approach allows for changes in views or shapes of the object with time. This flexibility is often absent in boosting approaches which match and boost the entire shape. However, boosting can be used with techniques such as multiple instance learning \cite{babenko2011robust} to allow for deformable object tracking.

\subsubsection{Optical flow based motion segmentation}\label{subsubsec:optical_Flow}

Optical flow methods have been used for motion segmentation as well \cite{mittal2004motion,ablavsky2003background,wixson2000detecting,vidal2004unified,cremers2005motion}. They incorporate spatial information by implying that the features or pixels with same motion characteristics are likely to belong to the same foreground object. Cues such as normalized color features \cite{mittal2004motion} may be used to augment foreground object detection. Some methods \cite{brox2006variational,cremers2005motion,amiaz2006piecewise} used partitioning of dense optical flow technique \cite{black1996robust} to deal with large motion variations. Videos are decomposed into different motion layers with distinct motion characteristics such that each layer has smooth motion characteristics and sharp motions appear only at the edges of motion layers. Level-set techniques are often found useful for computing the dense layers and their boundaries \cite{cremers2005motion}.

\subsubsection{Feature tracking and clustering for foreground tracking}

Another class of methods trace the features of the objects over the frames \cite{vidal2004unified,tron2007benchmark,sheikh2009background,brox2010object}. For example, sparse feature points are identified and tracked through out the video and then spectral \cite{brox2010object} or subspace \cite{vidal2010tutorial} clustering is applied to identify the clusters of features with same motion characteristics. Object segmentation is done by analysing the quality of clusters and post-processing \cite{ochs2011object}. We note that performing object segmentation after motion segmentation is different from performing tracking of features after object segmentation as discussed in section \ref{subsubsec:featuretracking}. Motion information is used for object segmentation in the former while object segmentation is used for extracting motion information in the latter.

Underlying assumptions in feature based motion segmentation are that the objects are rigid and the noise in data is limited to allow sparse tracking. While the first assumption is valid in maritime images for most scenarios, the second assumption may or may not be valid. It is notable that feature tracking methods are less computation intensive than optical flow approaches. Additionally, they can deal with random motion and large motion variations. Further, the features need not be pre-learnt. An arbitrarily large number of sparse features may be identified initially and only features with trackable motion may be retained. Other features may be classified as outliers and suppressed.

\subsubsection{Markov random field for foreground tracking}\label{subsubsec:MRF}

Zhou et. al \cite{zhou2013moving} modelled the motion of each individual outlier (which represents the foreground in the low rank background model) as a contiguous Markov random field. While the approach of \cite{zhou2013moving} is computationally elegant and performs background suppression, foreground segmentation, and motion segmentation simultaneously, Mumtaz  et. al \cite{mumtaz2014joint} reported that \cite{zhou2013moving} is not effective in suppressing wakes and shadows corresponding to the moving object. Thus, for making a method like \cite{zhou2013moving} more effective for maritime object problem, two approaches may be considered. The first approach is to augment the background model of \cite{zhou2013moving} with other background models such as those using local ternary patterns and visual saliency. The second approach is to use another model for background estimation, determine the corresponding support of the background (equivalent to the low rank matrix of \cite{zhou2013moving}) and then use the approach of \cite{zhou2013moving} to determine the outlier support and further motion modelling. Further, the estimation of the motion model may be considered as a predictive step and other suitable predictive models may be chosen if desired.

\subsubsection{Comparison of CV based tracking techniques} A comparative summary of tracking methods not covered in Table \ref{tab:tracking} is given in Table \ref{tab:comp_vis_foreground}. We compare performances of five computer vision techniques in Table \ref{tab:quant_track}. These techniques are AdaBoost (an online machine learning approach \cite{grabner2006real}), MIL (multi-instance learning, an online machine learning approach with support for multiple instance \cite{babenko2011robust}), TLD (tracking-learning-detection, a semi-supervised machine learning approach \cite{kalal2012tracking}), MedFlow (an optical flow technique, \cite{kalal2010forward}), and KCF (a color based feature tracking approach, \cite{danelljan2014adaptive}). The implementation in the openCV tracker library is used for these methods. In general, AdaBoost, Medflow, and KCF perform better than the other methods in the whole table. KCF performs the best in all metrics and is fast as well.

\vspace{-2mm}
\section{Concluding remarks}\label{sec:conclusion}

In this survey, contemporary works in maritime EO data processing have been discussed. For horizon detection, most of the work has been done by researchers working on maritime problems. In maritime EO data processing, object detection is done through segmentation of the foreground obtained after background subtraction. Thus, background subtraction is an important part of maritime EO data processing. Background subtraction may be performed on one image at a time assuming static background or may incorporate temporal information by modeling background as dynamic. In general, dynamic background approaches have better ability to deal with wakes, clouds, and foams. While a variety of methods are used in both categories of background subtraction, maritime EO data processing may benefit from other state-of-the-art background modelling techniques from the computer vision community as well. Different object tracking methods used in maritime EO processing are also discussed. Further, motion segmentation methods that do not segment the foreground to obtain objects but first learnt the motion patterns in the foreground and then cluster the patterns to identify objects are also discussed. Notably, while the gap between object tracking in maritime environment and computer vision community is relatively small, the gap in motion segmentation techniques is large. Nevertheless, we feel that motion segmentation may not be needed for on-board maritime processing since the scenes typically contain only a few objects of interest.

The study is supported with quantitative evaluation of performance of several representative maritime and computer vision techniques on Singapore Marine Dataset. This dataset has been created with the aim of providing challenging maritime EO videos for future research. The evaluation indicates that computer vision techniques can aid maritime vision with suitable advancement.

\appendix
\subsection{Postprocessing of maritime EO object tracking results}
Here, we discuss some useful post-processing of maritime detection or tracking results. Vessel's positions and speeds are more useful in physical units \cite{bloisi2009argos,withagen1999automatic}. Bloisi et. al \cite{bloisi2009argos} used high mounted stationary cameras such that the height of the vessels is irrelevant and all the water surface may be considered flat and comprising of only lateral coordinates in $xy$ plane. Further, the center of each pixel observed in the camera was mapped to a physical point through careful and extensive pre-acquisition calibration, which is possible owing to the fixed nature of the cameras. A tracking boat with differential global positioning system device (GPS) was used for calibration as well as position and velocity tracking test such that GPS accuracy of few cm and few cm/sec is achievable for position and velocity respectively. This indicates the difficulty in mapping tracking results to actual physical units and highlights the importance of augmenting tracking results with information from radar sensors. Nevertheless, for a given camera's fixed position and orientation, estimates of pixel to distance relationships may be obtained with limited accuracy and may help in providing comparative or fuzzy information about the speed and location of vessels, such as vessels far away, vessels approaching or receding, closer vessels, and fast moving vessels. A simplified physical distance mapping was proposed in \cite{withagen1999automatic} and is given as:
\begin{equation}\label{eq:physical_distance}
  d \approx \phi R - \sqrt{(\phi R)^2 - 2 hR}
\end{equation}
where $R$ is the radius of earth, $h$ is the height of camera, $\phi$ is the angle between the ship and horizon, and all of them are pre-known. Here, it is assumed that the object is at horizon. For a different point in the space, the angle between the point and the camera would be different. We note that this approximation is valid for points at far distances only.

It might be of interest to classify the vessels for their shape \cite{zhu2010novel}, size, speed, and visibility \cite{szpak2011maritime}. Such information may serve as indicator of the type of boat or vessel. Sometimes, crude or fuzzy classification such as very small (foe example, swimmer and debris), small (for example, jet ski, sail boat, and speed boat), medium (for example, fast boat, fishing boat, and steamers), and large vessels (for example, cruise ship and cargo ship) may suffice. Or, identification of the exact type and model of the vessel may be considered crucial in military or rescue scenarios and surveillance \cite{withagen1999automatic,van2012persistent}. There are two major approaches to classification and identification, which we discuss below.

The first is the approach of shape library, where segmentations of ground truth may be stored as references in the library and segmented shapes may be compared with the stored shapes to classify the segmented shapes \cite{frost2013detection}. The shape library must be sufficiently generative to be robust enough for reliable detection and sufficiently discriminative to be specific to the vessel type \cite{prasad2012survey}. For each vessel, shapes at different orientations \cite{van2008discriminating,van2012persistent} and spatial resolutions \cite{vicen2011ship} must be stored. Such approaches may require, in addition to the shape library, refined techniques for shape fitting, dominant point detections \cite{prasad2012novel}, and shape curvature analysis.

The other is a feature based approach in which a vessel is represented by a set of features discriminatively representing the vessel \cite{van2012persistent}. This approach requires selection of suitable features for shape classification \cite{prasad2012survey}. SIFT features \cite{van2012persistent,van2014ship,van2014recognition}, Haar features \cite{bloisi2011automatic}, Fisher vectors \cite{van2014ship,van2014recognition}, and statistical moments and their variations \cite{van2008discriminating,van2014ship,van2014recognition,zhu2010novel,gupta2009adaptive,van2012persistent} have been found useful for maritime vessels. It is argued in \cite{van2012persistent} that simple, though less discriminative, features such as localized moments from the electro-optic images may be sufficient when the object is at large distances and appears only a few pixels wide.

\vspace{-2mm}
\subsection{Multisensor approaches}\label{sec:examples_multimodal}

\begin{figure}
  \centering
  \includegraphics[width=0.7\linewidth]{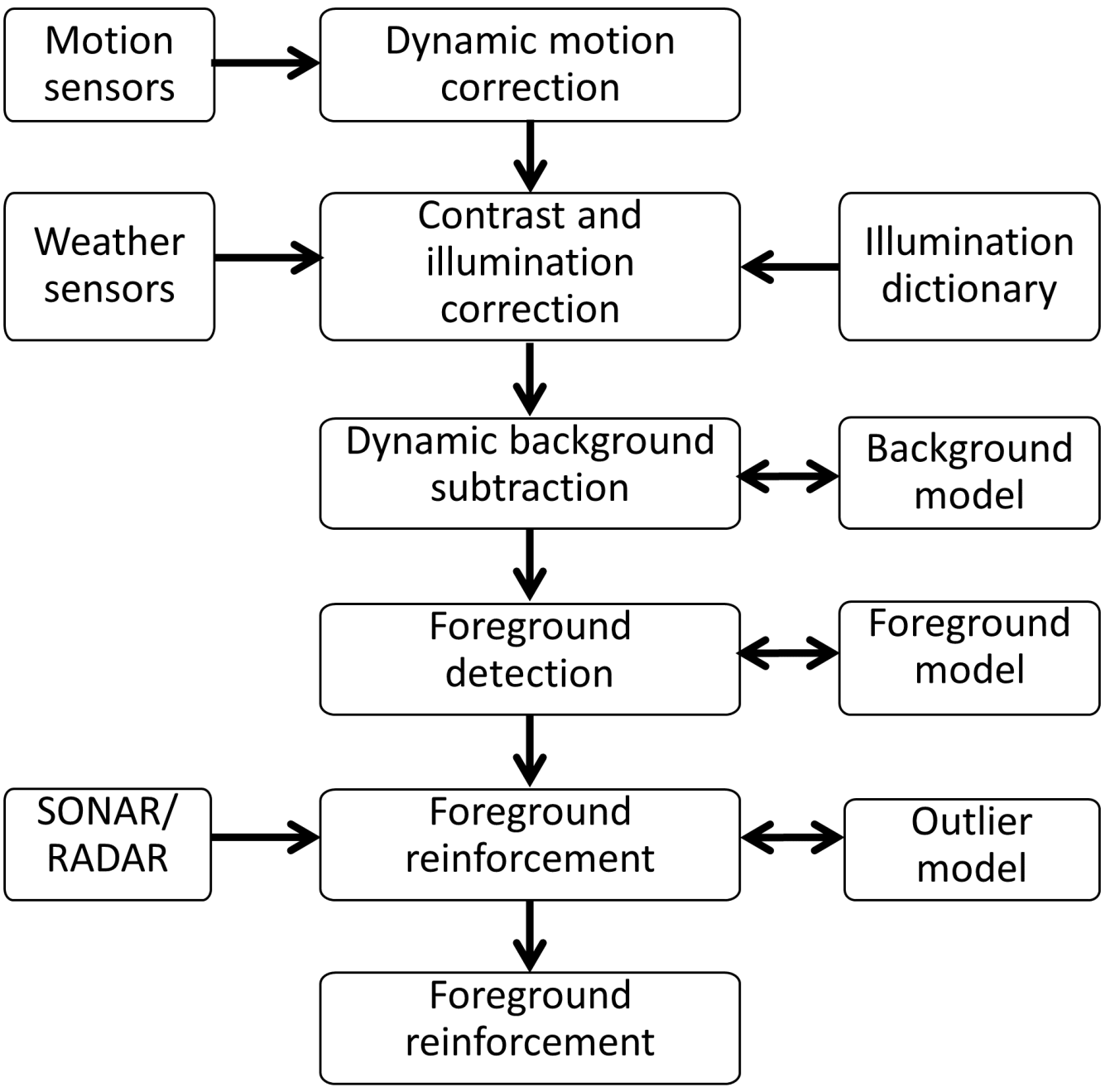}\vspace{-2mm}
  \caption{Flowchart of a multi-sensor processing system.}\label{fig:flowchart_multisensor}\vspace{-5mm}
\end{figure}

In this section, we review algorithms in which EO sensor data is processed in conjunction with other types of sensors, such as radar, sonar, gyroscope, and motion sensors. Although the focus of this survey is on EO sensor data processing alone, we believe it is useful to study the effectiveness of employing multiple sensors from other modalities to operate in tandem with EO sensors for tasks such as object detection and tracking. In particular, motion, gyro, and weather sensors can augment the EO data processing pipeline as shown in Fig. \ref{fig:flowchart_multisensor}. Also, radar and sonar can be used to filter outliers in EO data processing and vice versa \cite{bloisi2011automatic}.

In marine environment, \cite{babaee20153} combined electro-optical and sonar data stereoscopically to perform 3D reconstruction of floating objects. However, the work used a submerged electro-optical camera. Zheng et. al \cite{Zheng2008Mutlisensory} fused images from visible electro-optical sensor and IR sensor using discrete wavelet transform in an iterative fusion scheme to generate fused and pseudocolored images which have more information than individual sensors alone. Van den Broek et. al \cite{van2012persistent} presented a system architecture for multi-sensor data association. Specifically, radar or other spectral detectors were used to locate the ships and then zoomed-in images from visible or IR cameras were used to classify the ships based on pre-learnt discriminative feature libraries of the ships. Robert-Inacio et. al \cite{robert2007multispectral} combined data from a high definition color image and a 3-camera IR imaging system for video surveillance at a shore, with the particular intention of detecting terrorist threats. It showed that IR data has higher suppression of wake and thus enables better background detection, which was followed by high definition color data processing for threat analysis, as discussed in section \ref{subsubsec:basictracking}. Caspi and Irani \cite{caspi2002spatio} demonstrated that video sequences in a port scenario from a visible range and an IR sensor could be aligned by identifying video tracking features in each sequence and then finding point-to-point correspondences between the two sets of features. It may work well for images with similar spatial resolutions, but fail if the resolution scales of the two sensors are quite different.

Zhang et. al \cite{zhang2012visual} proposed to use rainfall radar, an in-situ multi-probe system equipped with turbidity, dissolved oxygen, temperature, conductivity and depth probes, and a visible sensor for a port-monitoring scenario. It reported that ships entering the port often coincide with spikes in data from the turbidity sensor, which are absent for small vessels. Schwering et. al \cite{schwering2007eo} discussed extensively the design requirements and challenges of an integrated electro-optic system which uses several multiband, hyperspectral, IR, visible range, as well as radar sensors to provide a reliable shore system with tracking, monitoring, and surveillance capabilities. Furthermore, an architecture of using EO sensor for maritime vehicle traffic system in populated areas is presented in \cite{bloisi2015bookchapter} and a practical account of maritime multi-sensor experiments is reported in \cite{elkins2010autonomous}.

\vspace{-2mm}
\subsection{Commercial maritime systems}\label{sec:patents}

Here, we discuss two commercial maritime systems, namely, a Vessel Identification and Positioning System (VIPS) and a patent on anti-collision warning system. Both use multisensor approach and the EO sensor is used as a part of larger scheme.

\subsubsection{Vessel Identification and Positioning System of Stratech Group Limited}
VIPS is an integrated on-shore sensor system developed by Stratech Group Limited\footnote{\url{http://www.thestratechgroup.com/iv_vips.asp}} for locating maritime vessels, estimating their heights and widths, and tracking them. The system uses electronic data from automatic identification system (AIS), radar, and EO sensors. The AIS and radar video provide information of potential locations of the vessels. Pre-calibrated electro-optical sensor system then zooms into the vicinity of coordinates provided by the AIS and the radar. It does so by segmenting small image regions around the coordinates. The segmented region is tracked as well as processed to derive height and width information \cite{chew2011method}.

\subsubsection{Anti-collision warning system for marine vehicle}
A patent \cite{waquet2010anti} approved in 2010 proposed a vessel-mounted anti-collision warning system which uses the EO sensors coupled to the compass as the main data source, which is augmented by AIS and radar for foreground reinforcement (see for example, the multi-sensor flowchart in Fig. \ref{fig:flowchart_multisensor}). The method first detects the horizon and looks for an object close to horizon. Once an object is located, it chooses a small image region around it and performs back ground subtraction using single image statistics (specifically, average intensity thresholding). Using the segmented shape, pre-calibrated EO-sensor grid, and the compass information, the azimuth of the object with reference to the vessel and its approximate height are computed. Also, an after-glow pattern (change in intensity of the segmented shape with time) is computed.
The temporal characteristics of the azimuth, the size, and the after-glow are compared with the reference visible objects' database and dangerous objects' database to determine whether an anti-collision warning should be generated. The comparison with the reference databases is done every 30 seconds and the history of an object is maintained for 20 minutes. The azimuth and size information can be checked against the AIS and radar system, or radar and AIS azimuthal tracks can be used instead of EO generated tracks.

\section*{Acknowledgement}
This work was conducted within Rolls Royce@NTU Corporate Lab with the support of National Research Foundation under the CorpLab@University scheme. The Singapore Marine Dataset is available at https://sites.google.com/site/dilipprasad/home/singapore-maritime-dataset.

\vspace{-2mm}
\bibliographystyle{IEEEtran}

\end{document}